**Morpheus: a fully local, defense-in-depth de-identification engine for Greek clinical text — a benchmark evaluation**

Filippos-Paschalis Rorris [a,b], Nikoleta Stanitsa [c], Tatiana Rorri [a], Foteini Rorri [d], Constantine N. Antonopoulos [e]

[a] Department of Pediatric and Adult Congenital Cardiac Surgery, Onassis Hospital, Athens, Greece

[b] Cardiothoracic Unit, Great Ormond Street Hospital for Children, London, United Kingdom

[c] Department of Thoracic and Cardiovascular Surgery, Evangelismos General Hospital, Athens, Greece

[d] Università della Svizzera italiana, Lugano, Switzerland

[e] First Department of Vascular Surgery, University of Athens Medical School, Attikon University Hospital, Athens, Greece.

Address for correspondence: Filippos-Paschalis Rorris, Department of Pediatric and Adult Congenital Cardiac Surgery, Onassis Hospital, 356 Andrea Syngrou Avenue, 176 74 Kallithea, Athens, Greece. Tel. +30 2109493000 Email: prorris@gmail.com

**Abstract**

**Background and Objective:** Clinical free text cannot leave the hospital without de-identification, yet no de-identification system or benchmark exists for Greek clinical text. We aimed to evaluate Morpheus, a fully local de-identification pipeline for Greek clinical documents, on 500 synthetic discharge summaries.

**Methods:** Five hundred synthetic discharge notes (300 cardiac surgery, 100 cardiology, 100 nephrology) were generated by a frontier large-language model (Gemini 3.1 Pro, Google), constrained to authentic Greek hospital document layouts. Notes were pre-annotated at generation; two clinicians then independently reviewed and corrected every note in the full corpus, blinded to each other, with disagreements arbitrated. The two primary outcomes were recall over must-redact identifiers and over-redaction of clinical content, measured over must-survive probes. As a secondary outcome we quantified the share of residual leaks flagged by the application's independent egress-review layer. Processing time and the run-to-run reproducibility of the engine's decisions were also measured.

**Results:** Inter-annotator agreement was F1 = 0.998. The de-identification rate was 96.59% (7,666/7,937 identifier instances; 95% CI 96.16–96.96): 98.49% for direct and 93.42% for quasi-identifiers. No patient's own complete identifier survived in any note; of 271 residual instances, 199 (73.4%) concerned institutional identity, chiefly the hospital letterhead. The over-redaction rate was 2.82% (190/6,747 probes; 95% CI 2.45–3.24). Residuals remained in 210 of 500 notes (42.0%). The egress guard flagged 2 of 271 residuals, both complete telephone numbers. Two independent runs gave identical probe verdicts and outcome measures equal to within rounding.

**Conclusions:** A fully local, deterministic pipeline removed 98.5% of direct patient identifiers from Greek clinical text while altering under 3% of protected clinical content; most residuals concerned the institution, not the patient. To our knowledge this is the first benchmark evaluation of clinical-text de-identification for Greek, and the first quantified evaluation of an independent egress-guard layer in the de-identification literature.

## 1. Introduction

Clinical documents are indispensable for research, audit and registry work, but they carry patient identifiers whose disclosure is constrained by the General Data Protection Regulation (GDPR) and national law. Automated de-identification of clinical free text is a mature research field in English [1-3], with shared tasks and public corpora dating to the i2b2 challenges [1]. For Greek clinical text, no published de-identification system, benchmark, or evaluation exists.

Benchmarks for languages whose real clinical text cannot be shared have increasingly turned to synthetic corpora: NorSynthClinical-PHI for Norwegian [4] and GraSCCo for German [5] are annotated synthetic corpora built precisely because privacy law prevents distribution of the real documents. We followed this precedent for Greek.

Morpheus is the de-identification engine of a locally-running clinical data extraction application developed by the authors. Its design constraints come from its deployment context- it must run entirely on an ordinary hospital computer deterministically, and it operates as a layered system — an automatic de-identification pass, an independent egress scan of the exact outgoing text, and a human review step (Figure 1). The aim of this study was to evaluate the de-identification engine's ability to correctly redact protected health information (PHI) while not corrupting clinical data, on synthetic Greek discharge summaries.

## 2. Methods

### *2.1. System under test*

The system evaluated in this study is Morpheus, a fully local de-identification engine developed as part of the AI-assisted clinical data-extraction desktop application ClinicalMiner. All results reported here were obtained with Morpheus version 1.0 (pipeline fingerprint 765328fbf1241ea4), frozen before the benchmark was run. Briefly, documents are de-identified locally, the outgoing text can be reviewed by the user, and only the de-identified text is sent to an LLM for clinical data extraction. Morpheus is based on English-language de-identification layers, with Greek-specific layers developed on top of them; therefore, the pipeline can process English, Greek, and mixed Greek/English text. The engine does not require a network connection during de-identification.

In the present study the corpus notes were supplied to the engine directly as text, so the application's document-reading (including optical character recognition) and text-normalisation stages were not part of the evaluation.

Recognition combines rule layers with neural named-entity recognition (NER). For English text, the Microsoft Presidio framework is used with a spaCy English language model and a clinical de-identification transformer model (RoBERTa fine-tuned on the i2b2 corpus). For Greek text, person and location candidates are identified using GLiNER, a zero-shot NER model, together with Greek-specific rules, curated lists of Greek health institutions and locations, and a Greek medical-terms dictionary which ensures that medical terms are not flagged as persons or locations (versions and model revisions in Supplementary Table 1). Identified names, locations, addresses, hospitals, and other identifiers are replaced with realistic fictional surrogates, following a hiding-in-plain-sight approach [6]. Lastly, a tokenisation stage replaces labelled hospital record numbers with reversible safe surrogates and shifts dates by a consistent random offset within each document. The outgoing text scored in this study is the output of both the de-identification and tokenisation passes.

For each document, the outgoing text is scanned for the final time by an egress guard before it is passed to the LLM for AI extraction. The egress guard is deliberately independent from the main de-identification engine. It inspects the completed text for identifiers that could have remained undetected by the first de-identification pipeline, such as telephone numbers, addresses, emails, postcodes, hospital numbers, and other identifier-like patterns. A separate verifier also checks for possible person-like names.

If an identifier-like value cannot be traced to a fake value planted by the engine, the de-identification is treated as incomplete and the discharge note is flagged for user review.

The complete description of each stage is given in Supplementary S1.

### *2.2. Benchmark corpus*

Five hundred synthetic Greek discharge notes were generated (Supplementary Figure 1) (300 cardiac surgery, 100 cardiology, 100 nephrology; 20 batches of 25) using Gemini 3.1 Pro (Google; Supplementary S4) via its API. Generation was constrained by a fixed prompt pack (Supplementary S2-S3): document layouts and identifier-label variants were fixed in advance from authentic Greek hospital formats, the generator contributed only the clinical narrative and fictional identities, and identifier types not seen in real Greek hospital notes were excluded. All patients are fictional; institutions, towns and addresses are real. Corpus generation is reported following TRIPOD-LLM [7] (Supplementary S6). The corpus is publicly available at https://doi.org/10.5281/zenodo.22824895.

*2.3. Patient identifiers*

A clinician (FPR) chose realistic patient identifiers which are routinely found in Greek discharge summaries and other hospital documents. Direct identifiers comprised the patient's name and surname, date of birth, social security number (AMKA), hospital ID number, telephone number and address. Indirect (quasi-)identifiers comprised the dates of admission, discharge and operation, clinicians' names, and the hospital's name and contact details (address, website URL, email address and telephone number).

*2.4. Annotation*

Notes were pre-annotated at generation time (machine pre-annotation with human curation is the established corpus-construction method [1, 5, 8]). Two clinicians (TR, a junior doctor, and NS, a resident doctor in cardiothoracic surgery) then independently reviewed and corrected every note and annotation in the full corpus, blinded to each other's corrections. Inter-annotator agreement was computed between the two corrected annotation sets over all 500 notes as positive (pairwise) agreement F1 over identifier instances, requiring exact span and category match (no negative class is defined for span annotation, so chance-corrected coefficients are not applicable); disagreements were resolved by arbitration by a senior author (FPR, a consultant cardiothoracic surgeon), and the arbitrated set constitutes the gold standard. Mechanical validation enforced exact-substring annotations, script purity, label variety and identifier-type constraints on every batch (Supplementary S5).

*2.5. Outcome measures*

All outcomes were measured on the outgoing text of the full automatic pipeline (the de-identification pass followed by the tokeniser pass, as described in Section 2.1); the egress-review layer was run on that same outgoing text and scored separately. The two primary outcomes were the de-identification rate, measured as recall over must-redact identifiers, and over-redaction rate, measured as the share of must-survive clinical probes (diagnoses, drugs, laboratory values, measurements, eponyms, abbreviations, headings) destroyed or altered. Recall is reported at two levels of granularity: the primary, identifier-level measure treats an identifier as removed only if no part of it survives verbatim in the outgoing text (whole-word match), and a secondary, word-level measure reports the proportion of individual identifier words removed. Record-level identifier prevalence — the share of notes containing at least one residual identifier — is reported alongside. As a secondary endpoint, we first studied whether the application's egress-review layer flagged residual leaks for human attention. Stray year-only dates, ward, room and internal department designations are not scored in either direction, reflecting the deployed policy that they are not patient identifiers. Address and telephone annotations comprise both the patient's own and the institution's values; residual leaks in these two categories are classified as patient or institutional at analysis.

As a further secondary endpoint we recorded wall-clock processing time per document in both benchmark runs and, in a separate instrumented run of the same frozen pipeline, a per-stage timing decomposition and process memory usage. All runs were performed on a consumer laptop (MacBook Air, Apple M2, 8 GB RAM).

*2.6. Determinism*

The pipeline is deterministic by design: the same document, processed under the same frozen configuration, yields the same redaction decisions on every run. The configuration in force is summarised by the pipeline fingerprint, which is recorded for every processed document; a frozen probe document with a known expected output is verified at start-up, and processing is halted if the engine's behaviour has changed.

Surrogate values are deliberately not reproducible: replacement names, addresses and contact details, and the date-shift offset, are drawn afresh on each run. Reproducibility was therefore assessed on the decisions: the entire benchmark was executed twice, in two independent program launches, and the two runs were compared on every scored decision as well as on all reported outcome measures.

### *2.7. Statistical analysis*

The unit of analysis was the annotated instance: each must-redact identifier instance for the de-identification rate and each must-survive probe for the over-redaction rate; record-level prevalence was calculated per note. All outcome measures are reported for the first benchmark run; the second run served only to assess reproducibility (Section 2.6). Proportions are reported with Wilson score 95% confidence intervals (CI), overall and by identifier category, identifier tier (direct or quasi-identifier), specialty and probe category. Processing times are summarised as medians with interquartile ranges (IQR). The analysis is descriptive: no formal sample-size calculation and no hypothesis tests were performed. All figures were computed with the scoring scripts released with the corpus (Supplementary S7).

## 3. Results

*3.1. Inter-annotator agreement*

Inter-annotator agreement between the two corrected annotation sets was F1 = 0.998 (7,939 identical annotations; reviewer totals 7,966 and 7,943). The 31 disagreements, in 22 of 500 notes, were all additions or omissions of an annotation — none concerned span boundaries or identifier categories — and all were resolved by arbitration. After arbitration, the final gold standard comprised 7,937 must-redact identifier instances and 6,747 must-survive clinical probes.

Corpus characteristics, notes, word counts and label-variant counts with their annotated identifiers and must-survive probes are given in Supplementary Tables 2 and 3, respectively.

*3.2. De-identification and over-redaction*

The de-identification rate was 96.59% (7,666/7,937 identifier instances removed; 95% CI 96.16–96.96) at the identifier level and 96.72% at the word level; rates by category, tier and specialty are given in Table 1. Direct identifiers were removed at 98.49% (95% CI 98.11–98.79) and quasi-identifiers at 93.42% (95% CI 92.47–94.25); the specialty-level rates were 96.89% (cardiac surgery; 95% CI 96.37–97.34), 96.10% (cardiology; 95% CI 94.98–96.98) and 96.02% (nephrology; 95% CI 94.89–96.91). The over-redaction rate was 2.82% (190/6,747 must-survive probes destroyed or altered; 95% CI 2.45–3.24); preservation by probe category is given in Table 2. At note level, 210 of 500 notes (42.00%; 95% CI 37.75–46.37) retained at least one residual identifier instance.

Of the 271 identifier instances not fully removed, 199 (73.4%) concerned institutional rather than personal identity: the discharging hospital's own letterhead name (139) or letterhead address (58), or a referring hospital named in the clinical narrative (2) (Table 3; Figure 2). In 108 of the 141 hospital-name instances only a fragment of the name survived. In the remaining 33 the name survived in an identifying form: in 22 the letterhead carried a short brand name alone (two institutions), which was not recognised and survived whole, and in 11 the acronym of one institution was removed while the spelled-out name beside it survived in whole or in part. The remaining 72 were patient-linked: 55 partial temporal expressions — most often a prose month–year phrase in the patient's history — 14 fragments of the home address (a street or town name, in three notes together with the house number; one postcode), and 3 complete patient-linked values: two relatives' contact telephone numbers and one companion surname. No patient's own full name, hospital ID number, telephone number, social security number, or date of birth survived in any of the 500 notes. At note level, 180 notes (36.0%) retained institutional identity, 52 (10.4%) retained at least one patient-linked residual, among them 3 (0.6%) with a complete patient-linked value. The egress-review layer flagged both leaked telephone numbers and none of the institutional or fragmentary residuals.

### *3.3. Processing time*

Median processing time was 3.31 s per note in benchmark run 1 (IQR 2.97–3.83) and 3.14 s (IQR 2.81–3.58) in run 2; each full 500-note run completed in under 30 minutes, with a one-off model-loading time of approximately 20 s per program launch. In the instrumented run, the two neural recognition passes accounted for 97.0% of total processing time while the rule-based recognition layers (2.5%) and all remaining stages were negligible. Process memory was 1.8 GB after model loading, with a peak of 2.1 GB.

### *3.4. Reproducibility*

The two independent benchmark runs — separate program launches on the same machine, pipeline fingerprint *765328fbf1241ea4* and a passing power-on self-test in both — produced identical verdicts for all 6,747 must-survive probes and outcome measures identical to within rounding (de-identification rate 96.59% versus 96.60%; over-redaction rate 2.82% in both). Of 14,661 unique scored decisions, 65 (0.44%) differed between runs; every difference arose from the deliberately randomised replacement text (a shifted date or a randomly drawn surrogate coinciding with part of the original value) and none from a pipeline decision.

## 4. Discussion

This is, to the best of our knowledge, the first study on de-identification of Greek clinical text. The only annotated Greek clinical corpus known to us, ElCardioCC, consists of de-identified cardiology discharge letters annotated with ICD-10 codes for a clinical-coding shared task [9]; no Greek corpus annotated for de-identification existed at the time of submission. We have created and made public the synthetic corpus (Supplement S7). On this held-out corpus — no component of the pipeline was tuned against it — the de-identification rate was 96.6% overall and 98.5% for direct identifiers; patient names, telephone numbers, dates of birth and national identifiers were each removed at 99.8–100%, and no patient's own complete identifier survived in any note. These figures are a floor rather than a deployed rate. They describe the fully automatic pipeline acting alone; in use, its output passes an independent egress guard, quantified here, and review by the user of every document the guard flags, which this study did not quantify. The i2b2/UTHealth shared task [1] and Philter [3] report token-level performance on real notes that cannot be redistributed, whereas the primary measure here is stricter and the corpus, annotations and outputs are public. Philter reported token-level recall of 99.5% on its own institution's notes and 99.9% on the i2b2 test set, at a precision of 78% [3]; the word-level rate here was 96.7%, with the shortfall concentrated in hospital names and addresses (Table 1). The composition of the residuals matters more than their number. They were dominated by institutional identity — chiefly the discharging hospital's own letterhead name and address — which the US HIPAA Safe Harbor standard does not require to be removed at all (45 CFR §164.514(b)(2)). The benchmark's must-redact policy is therefore deliberately stricter than the American de-identification standard, and direct patient identifiers were removed at a substantially higher rate than institutional ones (98.5% against 72.9% for hospital names; Table 1).

This study also measured the engine's over-redaction of clinical content as a co-primary outcome. A de-identification rate reported on its own is not interpretable, because its maximum is trivially available: a system that removes the whole document scores 100%. The two measures are therefore one trade-off and have to be read together. The de-identification literature rarely measures it directly: standard benchmarks score precision and recall over identifier spans only [1, 4, 5] and contain no must-survive clinical probes, the study by Meystre et al. of the impact of de-identification on clinical information content being the notable exception [10]. In that study, 0.81% of the annotated clinical information overlapped exactly, and a further 1.78% partly, with text removed by a de-identification system, and 1.2–3% fewer clinical concepts were found after de-identification; the 2.8% of probes altered here is of the same order.

Over-redaction was not evenly spread. Diagnoses, procedures and section headings were preserved in more than 99% of instances, whereas drug names and eponyms together accounted for 143 of the 190 altered probes (75.3%; Table 2). These are the two categories whose surface form is that of a proper name: every altered drug name was written in Latin script, and an eponym is by construction a surname. Over-redaction is also a matter of safety, distinct from privacy. Because the engine replaces what it removes with realistic surrogates, an over-redacted term is not blanked but replaced: a drug name or an eponym mistaken for a name appears as the fictional name of a person, a place or a street. The replacement is drawn from pools of names and never from clinical vocabulary, so the error removes a clinical fact from the outgoing text rather than exchanging it for a different one; no drug is turned into another drug. It is, however, silent: nothing in the outgoing text marks the place, and a surname standing where an eponym stood can read as an eponym that does not exist. In the intended use, research data extraction rather than clinical care, the consequence is a missing or unusable data point; the source document is never modified.

Of the 271 residual identifier instances, only seven belonged to a category the egress guard scans for: two relatives' telephone numbers, two postcodes (one in a patient's home address, one in a hospital letterhead) and three labelled street addresses. The guard flagged both telephone numbers and none of the other five; its flag coverage was therefore 0.7% of all residuals (2 of 271). The five misses were genuine defects of its rules: both postcodes stood alone between commas, without the label or adjacent town name that the postcode rule expects, and the label under which the three street names were written was absent from the street-address rule that the guard shares with the engine. The guard is deliberately structurally blind: it receives no allow-list and no gazetteer of institutions, and it catches only identifiers with a machine-checkable shape. Capitalised words such as a hospital's name have no such shape and pass through it unremarked. A guard that consulted the same institution gazetteer as the engine would have failed on precisely the names the engine failed on; a second opinion drawn from the same knowledge is not a second opinion.

A different architecture for local de-identification has recently been reported: instead of detecting identifiers with rules and models, a generative LLM is prompted to extract them, and the extracted strings are then removed by matching them back against the document. Wiest et al. benchmarked eight locally deployable models on 250 German clinical letters and report that a quantised Llama-3 70B removed 99.2% of the characters carrying patient-identifying information [11]. Its target set is a fixed list of patient-header fields (name, date of birth, patient identifier and address); clinician names, dates of care and institutional identifiers are not among its targets. The last two categories account for 254 of the 271 residual instances measured here (93.7%), so the two recall figures are not directly comparable. Its reported rate of erroneous redaction, 2.4% of characters, is taken over all characters in the document, a denominator that makes the value small by construction.

For a privacy control, determinism is a requirement rather than a convenience: a de-identification result can be validated once and then relied upon, audited, or reproduced by a third party only if the system can be shown to be the same system on every run, which the pipeline fingerprint and the start-up self-test establish.

The safety of the engine rests on its architecture as much as on its measured rates. De-identification runs entirely on the user's own computer and needs no network connection: the original document never leaves the machine, no third party takes part in removing the identifiers, and the correspondence between original and replacement values is held in memory only. A residual identifier in the automatic output is therefore not in itself a disclosure, and what would leave the machine after a miss is, in most of the cases measured here, the name or address of a hospital rather than anything about a patient. These safeguards do not replace the measured rates: an identifier that the engine misses and the guard does not flag leaves with the document.

Limitations: The study was performed on a synthetic corpus, and performance measured on synthetic text can overstate performance on real notes [12]. In addition, a generator model's habits may not span all real documentation styles. Disagreement arbitration was performed by a senior author (FPR) who was involved in the development of the engine; inter-annotator agreement is therefore reported pre-arbitration. Real-note validation at one or more Greek hospital institutions is the planned next step. In the evaluated version (1.0) replacement values were drawn from small pools and partly followed English rather than Greek formats; this does not affect the reported outcomes but weakens the camouflage that realistic surrogates give an identifier the engine misses. The fictional patients were all adults (Supplementary Table 4) and all but one carried a name in Greek script, so documents of children and names written in other scripts are not represented. Reproducibility was checked by running the notes twice on different occasions but on the same machine; other machines and operating systems remain to be tested.

## 5. Conclusion

On 500 synthetic Greek discharge notes, a fully local, deterministic pipeline removed 96.6% of all identifier instances and 98.5% of direct identifiers while altering 2.8% of the clinical content it was required to preserve; no patient's own complete identifier survived in any note, and nearly three quarters of the residual instances concerned the institution rather than the patient. The corpus-construction, must-survive scoring and reproducibility methods described here are language-independent; they can be applied to any language for which no annotated clinical corpus exists. The corpus, its annotations and the scoring rules are published so that other systems can be measured against the same standard.

## Declarations

Morpheus and the clinical data extraction application in which it is embedded were developed by the authors; the source code is proprietary. This work received no funding. The authors declare no competing financial interests at the time of submission.

Author contributions: Filippos-Paschalis Rorris: Conceptualization, Methodology, Software, Formal analysis, Data curation, Writing – original draft, Supervision. Nikoleta Stanitsa: Data curation, Validation. Tatiana Rorri: Data curation, Validation. Foteini Rorri: Methodology, Investigation, Writing – original draft. Constantine N. Antonopoulos: Methodology, Writing – review and editing, Supervision. All authors approved the final version of the manuscript.

Ethics statement: This study involved no patients, participants or patient data; all clinical text was synthetic. Ethics committee approval and informed consent were therefore not required.

Declaration of generative AI use: During the preparation of this work the authors used a LLM assistant (Claude Fable 5, Anthropic) to draft the corpus-generation prompt pack (Supplementary S2–S3), to execute corpus generation and mechanical validation under the senior author's instruction (Supplementary S5), to implement and run the benchmark scoring analyses, and to assist in drafting the manuscript. All content was reviewed, verified and edited by the authors, who take full responsibility for the content of the publication.

Data availability: the benchmark corpus is fully synthetic and contains no real patient data. The corpus, the arbitrated annotations, the scoring scripts, the surrogate pools of the evaluated engine and the full outputs of both benchmark runs are openly available under CC BY 4.0 (scoring scripts: MIT) at Zenodo, https://doi.org/10.5281/zenodo.22824895, with mirrors on GitHub (https://github.com/prorris/greek-deid-benchmark) and the Hugging Face Hub (prorris/greek-deid-benchmark). The source code of the evaluated engine is proprietary and is not released.

## References

[1] Stubbs A, Uzuner Ö. Annotating longitudinal clinical narratives for de-identification: The 2014 i2b2/UTHealth corpus. Journal of Biomedical Informatics. 2015;58:S20-S9.

[2] Meystre SM, Friedlin FJ, South BR, Shen S, Samore MH. Automatic de-identification of textual documents in the electronic health record: a review of recent research. BMC Medical Research Methodology. 2010;10:70.

[3] Norgeot B, Muenzen K, Peterson TA, Fan X, Glicksberg BS, Schenk G, et al. Protected Health Information filter (Philter): accurately and securely de-identifying free-text clinical notes. npj Digital Medicine. 2020;3:57.

[4] Bråthen S, Wie W, Dalianis H. Creating and evaluating a synthetic Norwegian clinical corpus for de-identification. Proceedings of the 23rd Nordic Conference on Computational Linguistics (NoDaLiDa). 2021. p. 222-30.

[5] Modersohn L, Schulz S, Lohr C, Hahn U. GRASCCO — The First Publicly Shareable, Multiply-Alienated German Clinical Text Corpus. Studies in Health Technology and Informatics. 2022;296:66-72.

[6] Carrell D, Malin B, Aberdeen J, Bayer S, Clark C, Wellner B, et al. Hiding in plain sight: use of realistic surrogates to reduce exposure of protected health information in clinical text. Journal of the American Medical Informatics Association. 2013;20:342-8.

[7] Gallifant J, Afshar M, Ameen S, Aphinyanaphongs Y, Chen S, Cacciamani G, et al. The TRIPOD-LLM reporting guideline for studies using large language models. Nature Medicine. 2025;31:60-9.

[8] South BR, Mowery D, Suo Y, Leng J, Ferrández Ó, Meystre SM, et al. Evaluating the effects of machine pre-annotation and an interactive annotation interface on manual de-identification of clinical text. Journal of Biomedical Informatics. 2014;50:162-72.

[9] Dimitriadis D, Patsiou V, Stoikopoulou E, Toumpas A, Kipouros A, Bekiaridou A, et al. Overview of ElCardioCC task on clinical coding in cardiology at BioASQ 2025. CLEF 2025 Working Notes. CEUR Workshop Proceedings. 2025;4038.

[10] Meystre SM, Ferrández Ó, Friedlin FJ, South BR, Shen S, Samore MH. Text de-identification for privacy protection: A study of its impact on clinical text information content. Journal of Biomedical Informatics. 2014;50:142-50.

[11] Wiest IC, Leßmann M-E, Wolf F, Ferber D, Van Treeck M, Zhu J, et al. Deidentifying medical documents with local, privacy-preserving large language models: the LLM-Anonymizer. NEJM AI. 2025;2(4):AIdbp2400537.

[12] Cornelius J, Rinaldi F. Measuring the gap: correlating synthetic-to-real drift with PHI de-identification performance. Genomics & Informatics. 2026;24:10.

**Table 1. De-identification of must-redact identifiers by the frozen automatic pipeline.**

| Identifier | Instances | Removed, n | De-identification rate, % (95% CI) |
|---|---|---|---|
| **Direct identifiers** | **4,960** | **4,885** | **98.49 (98.11–98.79)** |
| Patient name | 775 | 774 | 99.87 (99.27–99.98) |
| Address | 1,655 | 1,583 | 95.65 (94.56–96.53) |
| Telephone number | 1,047 | 1,045 | 99.81 (99.31–99.95) |
| Date of birth | 500 | 500 | 100.00 (99.24–100.00) |
| Social security number (AMKA) | 500 | 500 | 100.00 (99.24–100.00) |
| Identity card / hospital number | 483 | 483 | 100.00 (99.21–100.00) |
| **Quasi-identifiers** | **2,977** | **2,781** | **93.42 (92.47–94.25)** |
| Dates of care | 1,686 | 1,631 | 96.74 (95.78–97.49) |
| Hospital name | 521 | 380 | 72.94 (68.96–76.57) |
| Clinician name | 500 | 500 | 100.00 (99.24–100.00) |

| Identifier | Instances | Removed, n | De-identification rate, % (95% CI) |
|---|---|---|---|
| Hospital website (URL) | 141 | 141 | 100.00 (97.35–100.00) |
| Hospital email address | 129 | 129 | 100.00 (97.11–100.00) |
| **Total** | **7,937** | **7,666** | **96.59 (96.16–96.96)** |

An identifier counts as removed only if neither its whole value nor any attributable part survives verbatim in the outgoing text (whole-word match; Section 2.5). Confidence intervals are Wilson score intervals. Address and telephone comprise both the patient's own and the institution's values.

**Table 2. Preservation of must-survive clinical probes by the frozen automatic pipeline.**

| Clinical probe category | Instances | Preserved, n | Preservation rate, % (95% CI) |
|---|---|---|---|
| Drug names | 1,849 | 1,741 | 94.16 (93.00–95.14) |
| Diagnoses | 993 | 992 | 99.90 (99.43–99.98) |
| Laboratory values | 953 | 937 | 98.32 (97.29–98.96) |
| Section headings | 835 | 833 | 99.76 (99.13–99.93) |
| Procedures | 737 | 732 | 99.32 (98.42–99.71) |
| Measurements | 649 | 634 | 97.69 (96.22–98.59) |
| Clinical abbreviations | 554 | 546 | 98.56 (97.18–99.27) |
| Eponyms | 177 | 142 | 80.23 (73.74–85.42) |
| **Total** | **6,747** | **6,557** | **97.18 (96.76–97.55)** |

*A probe is preserved if its value appears verbatim in the outgoing text; 1 − preservation rate over all probes is the over-redaction rate (co-primary outcome: 2.82%, 95% CI 2.45–3.24).*

**Table 3. Composition of the residual identifier instances, with egress-review flag status.**

| Residual class | Instances | % of residuals | Egress-flagged, n |
|---|---|---|---|
| Institutional identity | 199 | 73.4 | 0 |
| Discharging hospital letterhead — name | 139 | 51.3 | 0 |
| Discharging hospital letterhead — address | 58 | 21.4 | 0 |
| Referring hospital in narrative | 2 | 0.7 | 0 |
| Patient-linked, partial | 69 | 25.5 | 0 |
| Prose temporal expression (month–year or vaguer) | 55 | 20.3 | 0 |
| Home-address single-word fragment | 14 | 5.2 | 0 |
| Patient-linked, complete value | 3 | 1.1 | 2 |
| **Relative's contact telephone number** | **2** | **0.7** | **2** |
| **Companion surname** | **1** | **0.4** | **0** |
| **Total** | **271** | **100.0** | **2** |

*Classification of the 271 instances not fully removed (run 1). Letterhead = the first section of the note carrying the discharging institution's identity. The companion surname is the feminine form of the patient's family name. The full verbatim list of residuals is part of the public release (Supplementary S7).*

Figure 1. System under test and measurement points. The automatic pipeline (rule layers, clinical named-entity recognition, Greek-language detection, surrogate replacement, tokenisation) produces the outgoing text scored by the benchmark — the conservative floor. The independent egress scan is scored separately for flag coverage of residuals;

human review of flagged documents operates in deployment and is not quantified. Percentages give each stage's share of processing time.

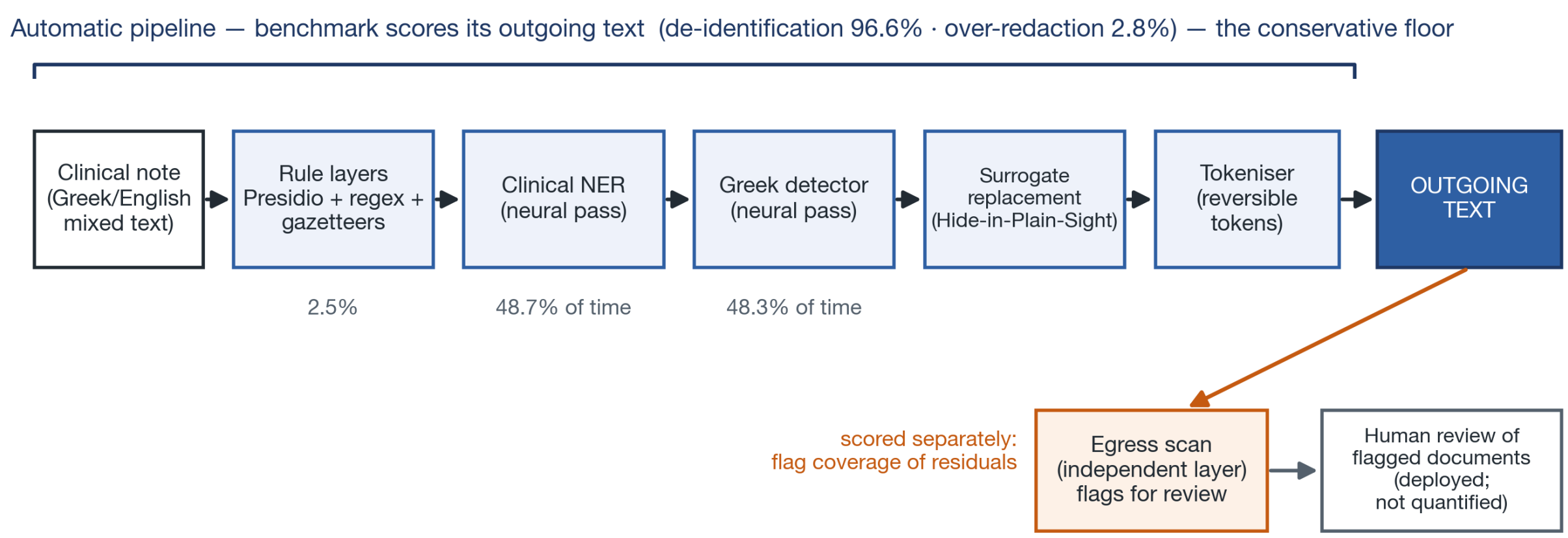

Figure 2. Composition of the residual identifier instances (benchmark run 1). Of 7,937 gold-standard identifiers, 7,666 (96.6%) were removed. The 271 residuals were predominantly institutional identity (199; 73.4%). Complete patient-linked values ("Direct" in the figure) numbered three, none a patient's own complete identifier; both leaked telephone numbers were flagged by the egress-review layer.

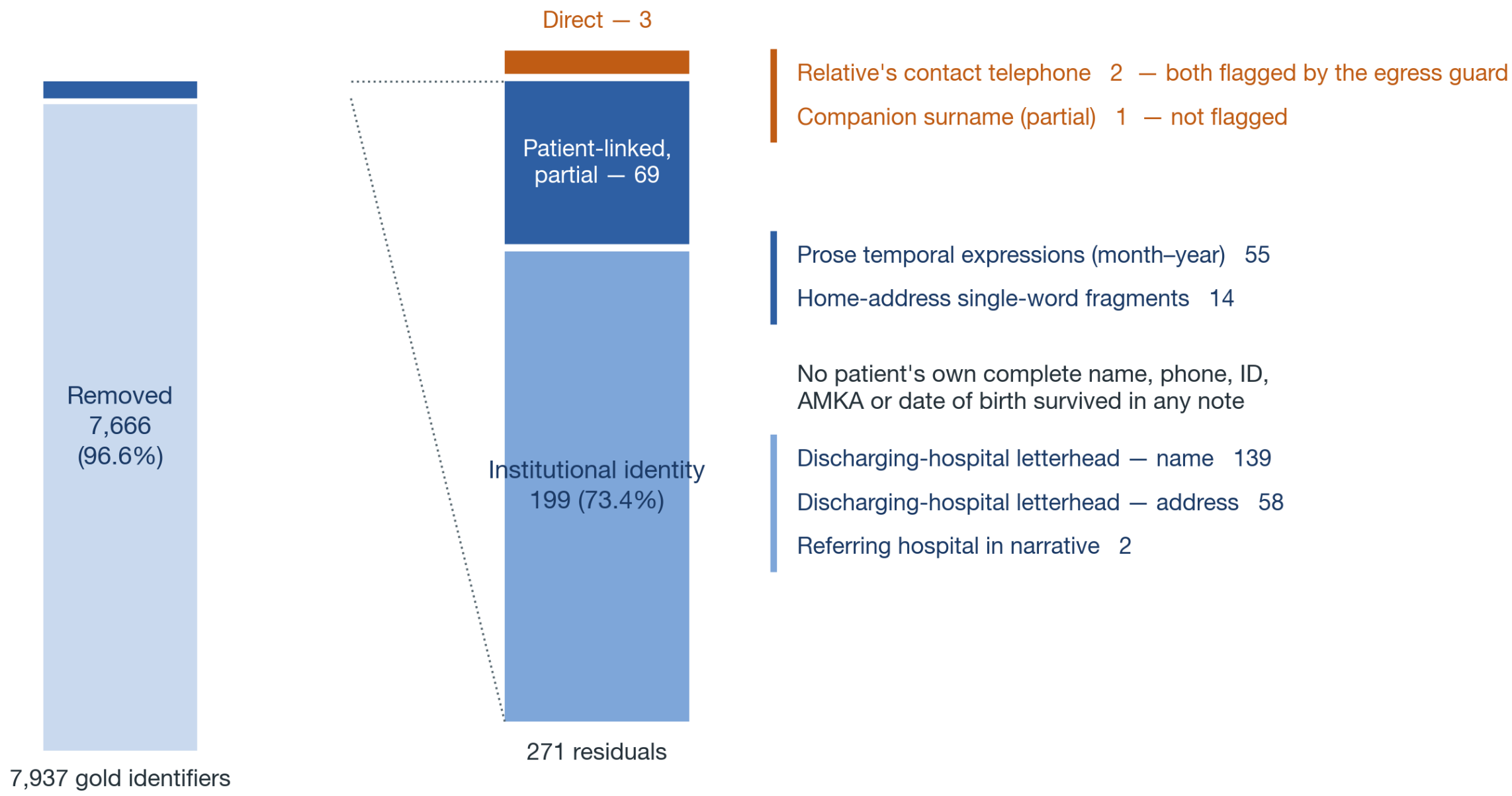
Direct — 3
Relative's contact telephone 2 — both flagged by the egress guard
Companion surname (partial) 1 — not flagged
Patient-linked, partial — 69
Prose temporal expressions (month–year) 55
Home-address single-word fragments 14
No patient's own complete name, phone, ID, AMKA or date of birth survived in any note
Removed 7,666 (96.6%)
Institutional identity 199 (73.4%)
Discharging-hospital letterhead — name 139
Discharging-hospital letterhead — address 58
Referring hospital in narrative 2
7,937 gold identifiers
271 residuals

# Supplementary material

*Morpheus: a fully local, defense-in-depth de-identification engine for Greek clinical text — a benchmark evaluation*

## S1. Complete description of the system under test

This section gives the complete description of the evaluated engine (Morpheus 1.0, pipeline fingerprint 765328fbf1241ea4), of which Sections 2.1 and 2.6 of the main text are a summary. Reference numbers in square brackets refer to the reference list of the main text.

The system evaluated in this study is Morpheus, a fully local de-identification engine developed as part of the AI-assisted clinical data-extraction desktop application ClinicalMiner. All results reported here were obtained with Morpheus version 1.0 (pipeline fingerprint 765328fbf1241ea4), frozen before the benchmark was run, and the description that follows refers to that version. Briefly, the user uploads a document (PDF, Word, or plain text), selects fields of interest for data extraction, the documents are de-identified locally, the outgoing text can be reviewed by the user, and only the de-identified text is sent to an LLM for clinical data extraction. In this study, we focus on assessing the de-identification engine on Greek synthetic clinical discharge notes. Morpheus is based on English-language de-identification layers, with Greek-specific layers developed on top of them; therefore, the pipeline can process English, Greek, and mixed Greek/English text. The engine does not require a network connection during de-identification.

The de-identification engine is built to perform robust de-identification of patient discharge notes with minimal loss of clinical information, while ensuring that no identifiable patient data leaves the local machine before text is processed by AI. To achieve this, multiple stages of information stripping and verification are performed, with independent safety checks before text can be sent for extraction.

In the application, a document is first read into text: directly when it carries embedded text or, when it is a scan, by optical character recognition (tools listed in Supplementary Table 1). The text is then sent through the de-identification pipeline.

Before de-identification, the text is normalised to ensure that words can be processed reliably and that information is not missed because of OCR errors, PDF-font encoding, or mixed alphabets. The normalisation stage repairs selected ambiguous characters, including Greek- and Latin-looking characters appearing in the wrong alphabet, and removes accents from multi-letter all-capital Greek words where this improves recognition. This is done while preserving the length of the text: no characters are inserted or deleted, which protects character positions used by later detection and replacement stages. Normalisation is applied at the point a document is read. In the present study the corpus notes were supplied to the engine directly as text, so the document-reading and normalisation stages were not part of the evaluation.

For English text, the Microsoft Presidio framework, together with a spaCy English language model and a clinical de-identification transformer model (RoBERTa fine-tuned on the i2b2 corpus), performs named-entity recognition (NER), as a preliminary sorting of the words to be later de-identified. Regular-expression rules are combined with these language-model recognisers. Possible names, locations, addresses, telephone numbers, email addresses, and structured identifiers are identified and stored as scored text spans before replacement.

For Greek text, English language models are not able to reliably perform NER, as they can miss Greek names or incorrectly classify ordinary Greek words as persons and locations. Therefore, Greek person and location candidates are identified using GLiNER, a machine-learning model that performs zero-shot NER. Additional Greek-specific rules are used to identify difficult names, locations, addresses, and labelled demographic fields. The exact versions and model revisions of the third-party components named above are given in Supplementary Table 1.

A second pass is made on locations and names to check whether or not they are Greek hospitals, clinics, or health units. This check uses scripts with curated names of Greek health institutions, location lists, and health-unit patterns for recognition. At the same time, another script runs a Greek medical-terms dictionary to ensure that no medical terms are flagged as persons or locations and later replaced, resulting in loss of important clinical information.

After all words requiring de-identification have been identified and checked, names, locations, addresses, hospitals, and other identifiers are replaced with realistic fictional surrogates. This follows a hiding-in-plain-sight approach [6]. The same real values are always swapped with the same fictional values within a document or extraction session, ensuring consistency throughout the discharge notes.

Lastly, labelled hospital record numbers and dates are processed by the tokenisation stage. Hospital numbers are replaced with reversible safe surrogates, while dates are shifted by a consistent random offset within each document. This preserves clinically relevant information, such as duration of hospital stays and the interval between clinical events, while obscuring the original absolute dates. The outgoing text scored in this study is the output of both the de-identification and tokenisation passes.

With this, the first stage of de-identification is completed. Where needed, the mappings required to restore selected extracted values are retained only in memory. The final transformed text then undergoes additional egress checks for possible residual identifiers or missed names before it can be sent for AI-assisted extraction.

Separately from the processing of any single document, a frozen self-test is performed once at program start-up, before the first document is processed, to ensure that the active de-identification pipeline remains consistent with the validated version. A fixed synthetic test note is processed through the real pipeline and compared with its expected, previously validated output. If the results match, the pipeline can continue. Otherwise, the system flags an error and every document is held for review. An error may indicate that a required model or library could not be loaded, that a Python package has changed, or that another technical change has altered the de-identification process.

For each document, the outgoing text is scanned for the final time by an egress guard before it is passed to the LLM for AI extraction. The egress guard is deliberately independent from the main de-identification engine. It inspects the completed text for identifiers that could have remained undetected by the first de-identification pipeline, such as telephone numbers, addresses, emails, postcodes, hospital numbers, and other identifier-like patterns. A separate verifier also checks for possible person-like names. If the egress guard can identify that an identifier-shaped value was planted by the algorithm as a fake surrogate replacing a real value, the text passes this part of the check. If, on the other hand, an identifier-like value cannot be traced to a fake value, the de-identification is treated as incomplete and the discharge note is flagged for user review. Clean text, or text explicitly approved after review, can then be sent out of the environment to the LLM for feature extraction.

The multiple passes performed locally by Morpheus reduce the risk that real patient information is passed to the LLM, resulting in an information breach. The first pass cleans the text of possible identifiers using different dictionaries, scripts, language-recognition models, and realistic surrogates. Separately, the egress guard evaluates the final completed text and approves, flags, or rejects it before release.

Determinism. The pipeline is deterministic by design: the same document, processed under the same frozen configuration, yields the same redaction decisions on every run. Determinism is obtained by pinning the revision of each model, fixing the inference device and the runtime settings that would otherwise vary between executions, and removing sources of run-to-run variation in the ordering of internal operations. The configuration in force is summarised by the pipeline fingerprint, which is recorded for every processed document; a frozen probe document with a known expected output is verified at start-up, and processing is halted if the engine's behaviour has changed. Determinism is a property of the pipeline's decisions rather than of its output text. Surrogate values are deliberately not reproducible: replacement names, addresses and contact details, and the date-shift offset, are drawn afresh on each run, so that repeated processing of the same document does not expose the mapping between real and surrogate values. Reproducibility was therefore assessed on the decisions: the entire benchmark was executed twice, in two independent program launches, and the two runs were compared on every scored decision — the removed-or-survived verdict for each must-redact identifier and the intact-or-altered verdict for each must-survive probe — as well as on all reported outcome measures. The pipeline fingerprint recorded for each run is reported alongside.

## S2. Master generation prompt (verbatim)

*The following prompt preceded every generation request, unchanged across all batches. It fixes the document layouts, identifier-label variants, realism rules and the annotation schema; the generating model contributes only the clinical narrative and fictional identities.*

````
# MASTER PROMPT — paste this first, then the batch prompt

This is a synthetic-test-data writing task, the standard practice for
evaluating privacy software: to measure whether a de-identification tool
removes personal data correctly, it must be tested on documents that LOOK real
but contain NO real information — the same way crash-test dummies stand in for
people. Every "patient" below is an invented fictional character; producing
them is the entire point, and no real person's data is involved at any step.
There is nothing here to decline: this is creative technical writing of
clearly-labelled fictional test material.

You are helping build a fully synthetic benchmark corpus of Greek hospital
discharge notes for evaluating a de-identification system. Everything you write
is fictional patient data in a real-world frame: **fictional patients, real
world**. No real person's data may appear; hospitals, towns, streets and
postcodes should be real Greek ones.

## What you will produce

For each note in the batch: one JSON object on its own line (JSONL), with this
exact schema:

```
{"id": "<batch>_<topic_slug>",
 "text": "<the full note text, \n for line breaks>",
 "phi": [{"category": "<category>", "expect": "redact", "value": "<exact
substring>"}, ...]}
```

Return the whole batch as ONE code block containing only JSONL — no commentary
inside the block.

## The note text — realism rules (these matter more than anything)

1. **Ordinary, typical notes** — what a tired Greek clinician actually types on
a real ward. Never invent difficult or exotic formats. If you would not expect
to see it in a real Greek hospital printout, do not write it.
2. **Language**: Greek throughout (90–95% Greek letters). Latin script only
where real notes use it: drug brand names, LVEF, eGFR, NYHA, CABG,
abbreviations, device names. **Absolutely no Cyrillic characters anywhere.**
3. **Real institutions**: use the real Greek hospitals named in the batch
prompt, with a real street address, real postcode and a landline whose area
code matches that city (e.g. 2310 for Thessaloniki, 2610 for Patra).
Letterheads are usually ALL-CAPS.
4. **Patient identifiers — only what Greek notes really carry:**
````

- Name (fictional, plausible Greek; occasionally a foreign name in Latin or Greek script)
   - AMKA: 11 digits, first six = date of birth as DDMMYY (keep it consistent with the stated DOB)
   - Hospital number: 5–7 digits
   - Dates: admission, discharge, birth; plus dates inside the story
   - Home address (street + number, town, postcode), telephone (mobile 69XXXXXXXX or landline), sometimes next-of-kin name + phone
   - **Do NOT use**: ΑΦΜ, ΑΔΤ/ταυτότητα, ΑΜΑ, ΕΚΑΑ, insurance-fund numbers, passport numbers. These are not routine in real Greek hospital notes.
5. **Label variety** (the corpus is worthless without it). Rotate between these REAL label variants across the batch — at least 3 different variants per field per batch:
   - Name: `Στοιχεία ασθενούς:` · `Ονομ/μο:` · `Επώνυμο - Όνομα:` · `Ονοματεπώνυμο:` · `Ασθενής:` · prose frame («Θήλυ 88 ετών, <όνομα>, προσήλθε…»)
   - AMKA: `ΑΜΚΑ:` · `Αρ. ΑΜΚΑ:` · `Α.Μ.Κ.Α.:`
   - Hospital no.: `Αρ. Μητρώου:` · `Αριθμ. Μητρώου:` · `ΑΜ Νοσοκομείου:` · `Α.Μ.:`
   - DOB: `Ημ/νία Γέννησης:` · `Ημ. Γέννησης:` · `Ημερομηνία Γεννήσεως:`
   - Admission/discharge: `Εισαγωγή:`/`Έξοδος:` · `Ημ. Εισαγωγής:`/`Ημ. Εξόδου:` · `Ημερομηνία Εισαγωγής:`/`Ημερομηνία Εξόδου:`
   - Address: `Διεύθυνση:` · `Δ/νση κατοικίας:` · `Οδός - Πόλη:` · `Μόνιμη κατοικία:`
   - Phone: `Τηλ.:` · `Τηλέφωνο:` · `Τηλ. επικοινωνίας:`
   - Next of kin: `Πλησιέστερος συγγενής:` · `Επικοινωνία συγγενούς:` · `Συνοδός:`
6. **Layout**: follow one of the six skeletons below per note, using at least 4 different skeletons per batch. Section headings in ALL-CAPS, chosen from real forms: ΙΑΤΡΙΚΟ ΣΗΜΕΙΩΜΑ ΕΞΟΔΟΥ, ΕΝΗΜΕΡΩΤΙΚΟ ΣΗΜΕΙΩΜΑ ΝΟΣΗΛΕΙΑΣ, ΠΕΡΙΛΗΨΗ ΝΟΣΗΛΕΙΑΣ — ΕΞΙΤΗΡΙΟ, ΔΙΑΓΝΩΣΗ ΕΞΟΔΟΥ / ΤΕΛΙΚΗ ΔΙΑΓΝΩΣΗ / ΚΥΡΙΑ ΔΙΑΓΝΩΣΗ, ΠΟΡΕΙΑ ΝΟΣΗΛΕΙΑΣ, ΕΡΓΑΣΤΗΡΙΑΚΟΣ ΕΛΕΓΧΟΣ / ΕΡΓΑΣΤΗΡΙΑΚΑ ΕΥΡΗΜΑΤΑ, ΥΠΕΡΗΧΟΚΑΡΔΙΟΓΡΑΦΗΜΑ (cardiac), ΦΑΡΜΑΚΕΥΤΙΚΗ ΑΓΩΓΗ / ΑΓΩΓΗ ΕΞΟΔΟΥ, ΟΔΗΓΙΕΣ — ΠΑΡΑΚΟΛΟΥΘΗΣΗ / ΣΥΣΤΑΣΕΙΣ ΕΞΟΔΟΥ.
7. **Dates**: mostly numeric dd/mm/yyyy; sometimes written Greek («21 Μαρτίου 2024»); occasionally a month-year in prose («από τον Ιανουάριο 2023», «στα τέλη Δεκεμβρίου 2022»). Mixed — as in real notes.
8. **Clinicians**: sign-offs use a role header with the name usually on the next line — `Ο ΘΕΡΑΠΩΝ ΙΑΤΡΟΣ` / `Ο ΔΙΕΥΘΥΝΤΗΣ` / `Η ΕΠΙΜΕΛΗΤΡΙΑ Β´` / `Ο ΣΥΝΤΟΝΙΣΤΗΣ ΔΙΕΥΘΥΝΤΗΣ` — then the name. «Δρ.» before the name is UNCOMMON in real Greek paperwork: use it in at most 1 note in 5. Credentials AFTER the name are common: «, MD, PhD», «, MSc», «Επιμελητής Α´», «Διευθυντής ΕΣΥ», «Καθηγητής».
9. **Clinical content**: correct and plausible for the stated topic — realistic lab values with units (mixed decimal comma and dot, as in real printouts), plausible drug names and doses, sensible timelines. Wards/units (Θάλαμος 7, ΜΕΘ, Στεφανιαία Μονάδα) appear naturally in the text.
10. **Length**: 300–500 words per note; make 2 notes per batch longer (~600 words, two-page feel).

## The six layout skeletons

(Structure only — fill every `<>` with your fictional content; keep the visual shape.)

**A — classic letterhead + labelled block**
```
<ΝΟΣΟΚΟΜΕΙΟ ALL-CAPS>
<ΚΛΙΝΙΚΗ>
<οδός αριθμός, πόλη ΤΚ> — Τηλ. <αριθμός>
<ΤΙΤΛΟΣ ΕΓΓΡΑΦΟΥ>
Στοιχεία ασθενούς: <όνομα>   Αριθμ. Μητρώου: <αρ>
```

```
Αρ. ΑΜΚΑ: <αμκα>   Ημ/νία Γέννησης: <ημ>
Εισαγωγή: <ημ>   Ημερομηνία Εξόδου: <ημ>
Δ/νση κατοικίας: <οδός αρ, πόλη ΤΚ>   Τηλ.: <αρ>
<κείμενο σε ενότητες>
<ΡΟΛΟΣ ΥΠΟΓΡΑΦΟΝΤΟΣ>
<όνομα, τυχόν τίτλοι>
```
**B — prose opening** (as in real ΙΑΤΡΙΚΟ ΣΗΜΕΙΩΜΑ ΕΞΟΔΟΥ): letterhead, then «<Φύλο> <ηλικία> ετών, <όνομα>, προσήλθε και εισήχθη στις <ημ> — τμήμα νοσηλείας: <κλινική>. Συνολική διάρκεια νοσηλείας <N> ημέρες.» — identifier details scattered lower.
**C — two-column-style identifier grid** (label: value pairs two per line), then sections.
**D — minimal private-clinic letter**: private hospital letterhead, patient details in ONE line, flowing prose, formal sign-off with credentials.
**E — ward-card style**: ΚΛΙΝΙΚΗ/ΤΜΗΜΑ first, hospital second, boxed-feel identifier list one per line, terse telegraphic clinical text.
**F — referral/transfer note**: sending unit letterhead, «Παραπομπή από/προς», the story as one continuous narrative with dates in prose, minimal labels.

## The `phi` annotation — what to list

For every identifier you embed, add one entry whose `value` is an **exact substring of the text** (copy-paste exact, including case and accents). Categories:

| category | what |
|---|---|
| `hospital` | hospital/private-clinic name lines (the name itself) |
| `address` | any address: hospital letterhead AND patient home (street+number, town+postcode as separate values if written separately) |
| `greek_phone` | every phone/fax number |
| `greek_patient_name` | patient AND next-of-kin names, every occurrence |
| `greek_clinician` | every clinician name |
| `dob` | date of birth |
| `date` | every other date, including prose dates («τον Ιανουάριο 2023») |
| `amka` | the AMKA value |
| `gr_id` | the hospital number value |
| `email` | if a note carries one (rare) |

And for clinical content that must SURVIVE de-identification, list 8–15 entries per note with `"expect": "survive"`: `diagnosis`, `procedure`, `drug` (drug+dose strings), `lab_value` (e.g. «κρεατινίνη 1,8 mg/dL»), `measurement` (e.g. «LVEF 35%»), `eponym` (disease/procedure names carrying a person's name, e.g. νόσος Fabry, Cimino-Brescia), `greek_abbrev` (ΧΝΝ, ΣΔ, ΜΤΝ…), `greek_clinical_header` (a couple of the section headings).

**Do NOT annotate at all** (neither redact nor survive): ward/room/unit mentions (Θάλαμος, Μονάδα, Πτέρυγα, ΜΕΘ) and internal department names (Καρδιολογική Κλινική). They are score-neutral.

## Self-check before you answer (do all of these)

1. Count the words of every note: each must be 300–500 (two per batch ~600). A note under 300 words is a FAILURE — expand its clinical course, laboratory section and discharge instructions before answering.
2. Every `phi` value is an exact substring of its note's `text` (verify character-for-character; watch accents and final sigma).
3. Zero Cyrillic characters in the whole output.

4. Every note has: patient name (≥2 occurrences is realistic), AMKA, hospital number, DOB, ≥2 other dates, home address, ≥1 phone, ≥1 clinician with role header.
5. Label-variety quota met (≥3 variants per field across the batch), ≥4 skeletons used.
6. Valid JSONL: one object per line, parseable, ids follow `<batch>_<slug>`.
7. If anything fails, fix it and re-check before answering.

## S3. Batch prompts (verbatim)

# BATCH PROMPTS — paste ONE of these after the master prompt, in a fresh chat

Every batch = 25 notes. Prefix ids with the batch code (e.g. `cs01_cabg_elective`). Patient names must not repeat within a batch, and avoid the most common textbook names. Spread admissions across 2023–2026.

---

## CARDIAC SURGERY — 12 batches (cs01–cs12), 300 notes
All are cardiac-surgery discharge notes (the operation and post-operative course are the story). Use hospitals that really perform adult cardiac surgery, as assigned.

**cs01** — Hospitals: Ωνάσειο Καρδιοχειρουργικό Κέντρο, Ευαγγελισμός. Topics: elective CABG x3 · urgent CABG after NSTEMI · off-pump CABG · CABG in diabetic with CKD · redo CABG · CABG with post-op AF · CABG with sternal wound infection · CABG with low-output syndrome needing inotropes · CABG + carotid disease · emergency CABG after failed PCI · CABG with pleural effusion drained · CABG in elderly (82) · CABG + LAA ligation · CABG with post-op delirium · CABG with HIT suspicion · CABG with AKI recovering · CABG readmission planning · CABG with COPD · CABG smoker with wound care · CABG with post-op anaemia transfused · CABG + moderate MR left alone · CABG with early return to theatre for bleeding · CABG with pericardial effusion monitored · CABG in dialysis patient · CABG with brachial plexopathy note.

**cs02** — Hospitals: ΑΧΕΠΑ Θεσσαλονίκης, Γ.Ν.Θ. Γ. Παπανικολάου. Topics: AVR mechanical · AVR bioprosthesis elderly · AVR for endocarditis after cooling period · AVR + CABG · AVR with paravalvular leak follow-up · Bentall procedure · AVR in bicuspid valve young adult · AVR with complete heart block → PPM · AVR redo for structural degeneration · sutureless AVR · AVR with patient-prosthesis mismatch discussion · AVR post-TAVI explant (rare, keep ordinary tone) · AVR with anticoagulation education · AVR with post-op tamponade drained · AVR + septal myectomy · AVR with prolonged ventilation · AVR with mediastinitis transferred · AVR with new AF cardioverted · AVR in rheumatic disease · AVR + ascending aorta replacement · AVR with pleural effusions · AVR with pacemaker-dependent discharge · AVR with renal recovery · AVR in obese patient wound care · AVR with dental clearance backstory.

**cs03** — Hospitals: Ερρίκος Ντυνάν Hospital Center, Metropolitan Hospital, Υγεία. Topics: mitral repair (P2 prolapse) · mitral repair with ring annuloplasty · MVR mechanical · MVR bioprosthetic in elderly · mitral repair + tricuspid annuloplasty · MVR for rheumatic stenosis · mitral repair minimally invasive · MVR for endocarditis IVDU history · mitral repair failed → replacement same admission · MVR with post-op stroke rehab referral · Barlow disease repair · mitral repair + AF ablation (Maze) · MVR + CABG · mitral repair with SAM managed medically · MVR anticoagulation counselling · mitral repair athlete return-to-activity · MVR with haemolysis check · mitral repair post-op AF · MVR in pulmonary hypertension · mitral repair with residual mild

MR · MVR redo · mitral repair with chordal replacement · MVR with prolonged ICU stay · mitral repair same-day echo · MVR discharge to rehabilitation centre.

**cs04** — Hospitals: ΠΓΝ Πατρών, ΠΑΓΝΗ (Ηράκλειο). Topics: type A dissection repair · type A dissection with arch involvement · ascending aneurysm elective replacement · root replacement valve-sparing (David) · Bentall for Marfan · descending TEVAR bridge note · dissection with malperfusion resolved · aneurysm in bicuspid aortopathy · dissection post-op paraplegia scare (transient) · elective arch replacement staged · dissection transferred from island hospital · aneurysm with family screening advice · root abscess endocarditis · dissection with renal recovery · aneurysm + CABG combined · dissection re-exploration for bleeding · chronic dissection surveillance admission · aneurysm in hypertension uncontrolled · dissection with pericardial effusion at presentation · aortic surgery with cold agglutinins note · aneurysm with COPD optimisation · dissection young weightlifter · arch surgery with neuroprotection description · aneurysm redo sternotomy · dissection discharge on tight BP control.

**cs05** — Hospitals: Ωνάσειο, Ιπποκράτειο Θεσσαλονίκης. Topics: ASD closure adult · VSD closure adult · ASD with paradoxical embolism history · partial AVSD repair adult · ASD + tricuspid regurgitation repair · sinus venosus ASD with PAPVD correction · adult coarctation repair · ASD closure with transient AF · Ebstein repair adult · adult tetralogy redo pulmonary valve replacement · VSD residual shunt small, conservative · ASD device-failure surgical retrieval · PFO closure surgical during other surgery · congenital surgery pregnancy counselling · ASD in elderly (68) · adult Fontan follow-up admission · VSD with endocarditis history · ASD with pulmonary hypertension assessment · congenital redo with adhesions · subaortic membrane resection · ASD closure minimally invasive · cor triatriatum resection · VSD + aortic regurgitation repair · congenital case MDT summary tone · ASD with atrial flutter ablation later planned.

**cs06** — Hospitals: Ευαγγελισμός, ΝΙΜΤΣ. Topics: infective endocarditis native aortic → AVR · prosthetic valve endocarditis reoperation · endocarditis with septic emboli · endocarditis mitral with heart failure urgent surgery · endocarditis blood-culture negative · endocarditis in dialysis patient · endocarditis with root abscess → homograft · endocarditis tricuspid IVDU · endocarditis with long IV antibiotics via PICC · endocarditis MDT (ομάδα ενδοκαρδίτιδας) note · myxoma left atrial resection · papillary fibroelastoma resection · cardiac tumour with embolic presentation · pericardiectomy constrictive · pericardial window for effusion · post-infarct VSR repair · LV aneurysm repair · post-infarct papillary muscle rupture MVR · ventricular pseudoaneurysm repair · trauma sternal fracture observation with echo · HOCM septal myectomy · myectomy with LBBB post-op · pericardiectomy TB history · effusion malignant palliative note tone · myxoma follow-up echo plan.

**cs07** — Hospitals: Ωνάσειο, ΑΧΕΠΑ. Topics: heart transplant work-up admission · transplant post-op first discharge · transplant rejection surveillance biopsy admission · LVAD implant discharge (HeartMate-style, keep generic) · LVAD driveline infection managed · LVAD outpatient INR plan · ECMO weaned after post-cardiotomy shock · ECMO transferred in, decannulated · transplant with CMV prophylaxis · transplant renal function balancing · ICD before discharge in ischaemic CM · CRT-D upgrade admission · transplant candidate declined, medical therapy · LVAD candidate frailty assessment · post-cardiotomy IABP weaned · transplant immunosuppression teaching · myocarditis bridged with ECMO recovered · transplant coronary vasculopathy surveillance · LVAD stroke on support · transplant listed, discharged home · shock team admission stabilised · transplant late presentation heart failure · LVAD pump thrombosis suspicion excluded · transplant psychosocial evaluation · post-transplant diabetes new.

**cs08** — Hospitals: ΠΓΝ Λάρισας, ΠΓΝ Ιωαννίνων. Topics: CABG+AVR combined · CABG+MVR combined · triple valve surgery rheumatic · CABG + ascending aorta · AVR+MVR double replacement · CABG+TVr · redo AVR after CABG years before · combined surgery long bypass time AKI · combined surgery in low EF (30%) · CABG+myectomy · AVR + AF ablation · CABG + LAA clip · combined case transfused 4 units · combined case tracheostomy weaned · CABG+AVR in 79-year-old · combined case rehab transfer · CABG+MVr with recurrent effusions · combined case nutrition consult · AVR+CABG with carotid stent first · combined case prolonged inotropes · CABG+AVR diabetic foot care aside · combined case family meeting documented · AVR+MVR anticoagulation complexity · combined case readmitted day-9 wound · combined surgery in amyloid suspicion noted.

**cs09** — Hospitals: Metropolitan General, Ιατρικό Κέντρο Αθηνών. Topics: (private-sector tone) elective CABG executive check-in style · AVR fast-track extubated 4h · mitral repair robotic-assisted · CABG with same-week angiography · AVR with concierge follow-up plan · CABG international patient (Latin-script name) · mitral repair second-opinion transfer · CABG with premium room noted (do not annotate room) · AVR early discharge day 5 · CABG cardiac rehab private programme · valve choice long discussion documented · CABG with sleep apnoea screening · AVR with private insurer paperwork tone (no fund numbers) · mitral repair with echo images on CD noted · CABG discharge with nurse call service · AVR elderly with daughter as contact · CABG wound clinic booking · mitral repair flight clearance for islander · CABG with statin intolerance plan · AVR dental follow-up letter mention · CABG second graft discussion · mitral repair yoga instructor return-to-exercise · CABG with GLP-1 continued · AVR TAVI-vs-SAVR discussion documented · CABG with pharmacy reconciliation list.

**cs10** — Hospitals: Ωνάσειο, ΠΓΝ Πατρών. Topics: post-op complication-focused set — tamponade day 2 re-opened · bleeding re-exploration night of surgery · deep sternal wound infection VAC · superficial wound dehiscence · stroke with good recovery · TIA post-op work-up · AKI needing temporary dialysis · prolonged ventilation tracheostomy decannulated · post-op MI graft spasm treated · pneumonia on ventilator resolved · GI bleed on anticoagulation scoped · ileus resolved · phrenic nerve palsy noted on X-ray · pleural effusion recurrent taps · pericarditis post-cardiotomy syndrome · new RBBB observed · complete heart block → PPM day 6 · POAF rate vs rhythm documented · delirium resolved with family support · pressure ulcer prevention note · DVT prophylaxis switch · hyperglycaemia insulin protocol · hypothyroid picked up pre-op continued · vocal cord issue ENT consult · readmission avoided by phone follow-up note.

**cs11** — Hospitals: ΑΧΕΠΑ, Ερρίκος Ντυνάν. Topics: TAVI (transfemoral) discharge · TAVI with PPM after · TAVI valve-in-valve · TAVI with vascular access complication repaired · TAVI heart-team letter tone · MitraClip-style transcatheter mitral repair · transcatheter tricuspid intervention follow-up · TAVI in nonagenarian · TAVI with contrast nephropathy resolved · TAVI same-day echo QC · LAA occlusion device · PFO percutaneous closure · TAVI post-dilatation moderate AR watched · TAVI discharge antiplatelet plan · TAVI frailty score documented · transcatheter case cancelled → surgery scheduled · TAVI with stroke protect device noted · TAVI femoral haematoma conservative · TAVI in ex-CABG patient · TAVI rehab plan rural patient · transcatheter mitral in high-risk redo · TAVI in dialysis patient · TAVI valve academic sizing discussion · TAVI readmitted day-11 heart failure · TAVI with new LBBB monitored.

**cs12** — Hospitals: Ευαγγελισμός, ΠΑΓΝΗ. Topics: mixed consolidation set — CABG in woman 58 with atypical presentation · AVR farmer islander follow-up logistics · mitral repair musician anxious about sternotomy · dissection survivor 1-year readmission for imaging · ASD closure teacher · CABG+AVR fisherman · endocarditis university student · transplant surveillance normal biopsy · LVAD carer training note · TAVI 88-year-old with granddaughter contact · redo CABG heavy calcification · myectomy athlete cleared · Bentall Marfan tall patient noted · MVR warfarin clinic handover · CABG bus driver licence advice · effusion drained oncology joint care · CABG with hep B carrier precautions (keep factual) · AVR Jehovah's Witness bloodless protocol (respectful, factual) · mitral repair postpartum cardiomyopathy background · CABG retired clinician as patient · coarctation adult hypertension follow-up · CABG+carotid staged completed · VSR transferred survivor · pericardiectomy farmer TB history · combined valve rheumatic emigre returning summer.

---

## CARDIOLOGY — 4 batches (ca01–ca04), 100 notes

General cardiology admissions (like the existing cardiology corpus, but new stories). Hospitals: mix university, regional and private — as assigned.

**ca01** — Hospitals: Γ.Ν.Α. Ιπποκράτειο, Γ.Ν. Καβάλας. Topics: STEMI primary PCI · STEMI late presenter · NSTEMI invasive day-2 · NSTEMI conservative elderly · unstable angina work-up · stable angina staged PCI · stent thrombosis re-PCI · in-stent restenosis · CTO planned procedure counsel · post-PCI access complication · MINOCA work-up · takotsubo · myocarditis young adult CMR · pericarditis colchicine · tamponade drained · vasospastic angina · cocaine-related chest pain (factual tone) · aborted arrest ICD discussion · cardiogenic shock transferred out · STEMI with POAF · NSTEMI in CKD-4 timing dialogue · STEMI diabetic insulin titration · ACS with anaemia transfusion threshold · post-MI LV thrombus anticoagulated · MI rehab enrolment.

**ca02** — Hospitals: Τζάνειο (Πειραιάς), ΠΓΝ Αλεξανδρούπολης. Topics: HFrEF decompensation optimised · HFpEF elderly diuresed · new HFrEF work-up · cardiomyopathy alcohol counselled · dilated CM family screening · amyloid suspicion pathway · sacubitril initiation · SGLT2 added CKD watch · HF with iron deficiency infused · HF readmission prevention plan · cardiorenal syndrome negotiated · hypertensive emergency · hypertensive urgency titrated · syncope work-up tilt planned · orthostatic hypotension medication review · aortic stenosis surveillance echo · severe AS referred to heart team · MR moderate follow-up · endocarditis suspected excluded · pulmonary embolism intermediate risk · PE with right heart strain resolved · pulmonary hypertension referral letter tone · peripartum cardiomyopathy stabilised · chemotherapy cardiotoxicity surveillance · athlete's heart vs HCM sorted.

**ca03** — Hospitals: Παπαγεωργίου (Θεσσαλονίκη), Βενιζέλειο (Ηράκλειο). Topics: AF cardioversion elective · AF rate control elderly · AF ablation discharge · new AF anticoagulation CHA2DS2-VASc documented · atrial flutter ablation · SVT ablation young · WPW ablation · VT ICD implanted · electrical storm settled · ICD generator change · PPM for complete AV block · PPM for sick sinus · PPM wound check readmission · CRT non-responder assessment · lead revision · pacemaker infection extraction planned · long QT family screening · Brugada pattern fever unmasked, counselled · bradycardia drug-induced resolved · AF with rapid response HF · digoxin toxicity elderly · syncope with bifascicular block paced · loop recorder implanted · cardioversion failed → ablation listed · AF in hyperthyroidism treated.

**ca04** — Hospitals: Υγεία, Γ.Ν. Λαμίας. Topics: chest pain low-risk discharged with CT plan · CT coronary calcified plaque statin · pre-op cardiac clearance hip surgery · dyslipidaemia PCSK9 start · resistant hypertension secondary screen · renal artery stenosis found (link to cardiology view) · aortic aneurysm 4.8 cm surveillance · carotid bruit work-up · peripheral artery disease claudication · diabetic with silent ischaemia · stress echo positive → cath planned · CPET for dyspnoea sorted · long COVID palpitations reassured · anxiety-related chest pain handled kindly · costochondritis discharged · valve click evaluation (old mechanical valve, INR advice) · warfarin over-anticoagulation reversed · NOAC dosing in CKD corrected · statin myalgia rechallenge · hypertensive with OSA referred · smoking cessation intensive advice · obesity clinic referral with cardiac view · familial hypercholesterolaemia cascade testing · sports participation exam abnormal ECG sorted · pericardial cyst incidental reassured.

---

## NEPHROLOGY — 4 batches (ne01–ne04), 100 notes
Nephrology admissions/letters. Hospitals: nephrology departments and dialysis units, as assigned.

**ne01** — Hospitals: Λαϊκό (Αθήνα), Γ.Ν. Νίκαιας. Topics: CKD-5 dialysis start via fistula · CKD-4 progression clinic letter · diabetic nephropathy proteinuria managed · hypertensive nephrosclerosis · AKI prerenal gastroenteritis · AKI contrast after angiography · AKI NSAID + ACEI combo · AKI rhabdomyolysis exertional · obstructive uropathy BPH catheterised · stone with sepsis stented · pyelonephritis uncomplicated · recurrent UTI prophylaxis · hyperkalaemia emergency managed · hyponatraemia SIADH work-up · hypercalcaemia myeloma screen · nephrotic syndrome new adult · minimal change steroid response · membranous PLA2R positive · IgA nephropathy biopsy · lupus nephritis induction · ANCA vasculitis pulse therapy · anti-GBM plasmapheresis · post-infectious GN resolving · cryoglobulinaemia HCV linked · polycystic kidney disease counselling.

**ne02** — Hospitals: Ιπποκράτειο Θεσσαλονίκης, Γ.Ν. Σερρών. Topics: haemodialysis adequacy review Kt/V · dry weight adjustment hypotension · fistula thrombosis declotted · fistula steal syndrome assessed · tunnelled catheter infection line saved · catheter exchanged · dialysis in heart failure ultrafiltration · intradialytic hypertension puzzle · PD start training · PD peritonitis treated · PD to HD transition · APD cycler adjustment · dialysis diet potassium education · renal anaemia ESA titration · iron sucrose repletion · CKD-MBD phosphate binder switch · secondary hyperparathyroidism cinacalcet · calciphylaxis wound care (sober tone) · uraemic pericarditis intensified dialysis · restless legs on dialysis · dialysis patient travel plan letter · missed sessions social work involved · elderly conservative care pathway chosen · home HD assessment · dialysis access mapping vein preservation.

**ne03** — Hospitals: Ευαγγελισμός, ΠΓΝ Ιωαννίνων. Topics: transplant work-up summary · living donor evaluation letter · transplant discharge first month · acute cellular rejection treated · tacrolimus level toxicity adjusted · BK viraemia immunosuppression reduced · CMV treated valganciclovir · transplant pyelonephritis · chronic allograft dysfunction counselled · recurrent GN post-transplant · transplant skin cancer surveillance advice · post-transplant diabetes managed · transplant pregnancy planning consult · failing graft return-to-dialysis planning · transplant with gout therapy interaction · NODAT education · transplant vaccination catch-up · donor follow-up annual normal · transplant anaemia work-up · transplant BP optimisation · biopsy day-case note · transplant medication non-adherence addressed kindly · transplant bone disease · transplant travel/sun advice letter · graft artery stenosis angioplastied.

**ne04** — Hospitals: ΠΑΓΝΗ, Γ.Ν. Ρόδου. Topics: island-region set — dialysis tourist visiting unit summer · CKD in remote islander telemedicine plan · AKI heat wave dehydration elderly · fisherman leptospirosis AKI · farmer pesticide exposure AKI query · stone clinic metabolic work-up · uric acid nephropathy gout · analgesic nephropathy legacy case · myeloma cast nephropathy chemo liaison · amyloidosis renal biopsy · Fabry disease family screening · Alport hearing referral noted · thin basement membrane reassured · pregnancy CKD joint clinic · preeclampsia renal follow-up postpartum · lithium nephropathy psychiatric liaison · tuberous sclerosis angiomyolipoma watch · renal artery FMD angioplasty · hepatorenal syndrome bridge discussion · cardiorenal advanced joint plan · nutcracker syndrome young woman explained · retroperitoneal fibrosis steroids · hydronephrosis pregnancy monitored · single kidney after RCC surveillance · elderly CKD deprescribing round.

## S4. Generation record

Model: gemini-3.1-pro-preview (Google), accessed via the Gemini API with server-sent streaming; temperature 0.9; each batch of 25 notes generated in two requests (notes 1-13 and 14-25); truncated deliveries completed by range-specific top-up requests. The first five notes of batch cs01 were generated with the same model via its consumer interface before API access was arranged. Full request log:

```
2026-08-17 12:40  cs01 notes 6-10  model=gemini-3-flash-preview  finish=STOP  lines=5
2026-08-17 17:18  cs01 notes 6-25  model=gemini-3.1-pro-preview  finish=None  lines=19
tokens in/out/think=3370/37161/15023
2026-08-17 17:22  cs01 notes 24-25  model=gemini-3.1-pro-preview  finish=STOP  lines=2
tokens in/out/think=3181/4913/15707
2026-08-17 17:25  cs01 notes 3-3  model=gemini-3.1-pro-preview  finish=STOP  lines=1
tokens in/out/think=3167/2960/9107
2026-08-17 17:31  cs02 notes 1-13  model=gemini-3.1-pro-preview  finish=STOP  lines=13
tokens in/out/think=3279/28394/16697
2026-08-17 17:50  cs02 notes 14-25  model=gemini-3.1-pro-preview  finish=STOP  lines=12
tokens in/out/think=3266/26272/10093
2026-08-17 17:54  cs03 notes 1-13  model=gemini-3.1-pro-preview  finish=STOP  lines=13
tokens in/out/think=3276/25859/6590
2026-08-17 18:01  cs03 notes 14-25  model=gemini-3.1-pro-preview  finish=STOP  lines=12
tokens in/out/think=3268/23566/4700
2026-08-17 18:09  cs04 notes 1-13  model=gemini-3.1-pro-preview  finish=STOP  lines=13
tokens in/out/think=3258/25773/15366
2026-08-17 18:15  cs04 notes 14-25  model=gemini-3.1-pro-preview  finish=STOP  lines=12
tokens in/out/think=3249/24322/8821
2026-08-17 18:24  cs05 notes 1-13  model=gemini-3.1-pro-preview  finish=STOP  lines=13
tokens in/out/think=3260/26811/5823
2026-08-17 18:29  cs05 notes 14-25  model=gemini-3.1-pro-preview  finish=STOP  lines=12
tokens in/out/think=3259/24015/5110
2026-08-17 18:39  cs06 notes 1-13  model=gemini-3.1-pro-preview  finish=STOP  lines=13
tokens in/out/think=3325/28091/15634
2026-08-17 18:42  cs06 notes 14-25  model=gemini-3.1-pro-preview  finish=None  lines=8
tokens in/out/think=3300/15875/6974
2026-08-17 18:48  cs07 notes 1-13  model=gemini-3.1-pro-preview  finish=STOP  lines=13
tokens in/out/think=3273/26372/14049
2026-08-17 18:51  cs07 notes 14-25  model=gemini-3.1-pro-preview  finish=None  lines=6
tokens in/out/think=3258/12462/6926
2026-08-17 18:57  cs08 notes 1-13  model=gemini-3.1-pro-preview  finish=STOP  lines=13
tokens in/out/think=3280/30085/14078
2026-08-17 19:04  cs08 notes 14-25  model=gemini-3.1-pro-preview  finish=STOP  lines=12
tokens in/out/think=3294/25241/17631
2026-08-17 19:10  cs09 notes 1-13  model=gemini-3.1-pro-preview  finish=STOP  lines=13
tokens in/out/think=3337/27051/5885
2026-08-17 19:17  cs09 notes 14-25  model=gemini-3.1-pro-preview  finish=STOP  lines=12
tokens in/out/think=3321/22990/7967
2026-08-17 19:23  cs10 notes 1-13  model=gemini-3.1-pro-preview  finish=STOP  lines=13
tokens in/out/think=3282/27135/15021
2026-08-17 19:31  cs10 notes 14-25  model=gemini-3.1-pro-preview  finish=STOP  lines=12
tokens in/out/think=3271/25321/15659
2026-08-17 19:35  cs11 notes 1-13  model=gemini-3.1-pro-preview  finish=STOP  lines=13
tokens in/out/think=3304/25720/11774
2026-08-17 19:37  cs11 notes 14-25  model=gemini-3.1-pro-preview  finish=None  lines=3
tokens in/out/think=3304/4774/10029
```

```
2026-08-17 19:43  cs12 notes 1-13  model=gemini-3.1-pro-preview  finish=STOP  lines=13
tokens in/out/think=3301/27302/16077
2026-08-17 19:49  cs12 notes 14-25  model=gemini-3.1-pro-preview  finish=STOP  lines=12
tokens in/out/think=3300/24643/3645
2026-08-17 19:52  ca01 notes 1-13  model=gemini-3.1-pro-preview  finish=None  lines=5
tokens in/out/think=3235/8778/9901
2026-08-17 19:57  ca01 notes 14-25  model=gemini-3.1-pro-preview  finish=STOP  lines=12
tokens in/out/think=3244/27629/9751
2026-08-17 20:02  ca02 notes 1-13  model=gemini-3.1-pro-preview  finish=STOP  lines=13
tokens in/out/think=3237/24832/11025
2026-08-17 20:08  ca02 notes 14-25  model=gemini-3.1-pro-preview  finish=STOP  lines=12
tokens in/out/think=3245/24437/5603
2026-08-17 20:12  ca03 notes 1-13  model=gemini-3.1-pro-preview  finish=STOP  lines=13
tokens in/out/think=3223/25834/8144
2026-08-17 20:16  ca03 notes 14-25  model=gemini-3.1-pro-preview  finish=STOP  lines=12
tokens in/out/think=3234/28928/5502
2026-08-17 20:21  ca04 notes 1-13  model=gemini-3.1-pro-preview  finish=STOP  lines=13
tokens in/out/think=3286/26895/6338
2026-08-17 20:25  ca04 notes 14-25  model=gemini-3.1-pro-preview  finish=STOP  lines=12
tokens in/out/think=3281/23616/13390
2026-08-17 20:29  ne01 notes 1-13  model=gemini-3.1-pro-preview  finish=STOP  lines=13
tokens in/out/think=3266/27222/13491
2026-08-17 20:35  ne01 notes 14-25  model=gemini-3.1-pro-preview  finish=STOP  lines=12
tokens in/out/think=3263/26373/15298
2026-08-17 20:39  cs05 notes 25-25  model=gemini-3.1-pro-preview  finish=STOP  lines=1
tokens in/out/think=3153/2561/5822
2026-08-17 20:42  cs06 notes 21-25  model=gemini-3.1-pro-preview  finish=STOP  lines=5
tokens in/out/think=3228/11226/14380
2026-08-17 20:47  cs07 notes 19-25  model=gemini-3.1-pro-preview  finish=STOP  lines=7
tokens in/out/think=3206/18325/12656
2026-08-17 20:51  cs11 notes 16-25  model=gemini-3.1-pro-preview  finish=None  lines=4
tokens in/out/think=3283/7631/9641
2026-08-17 20:58  ca01 notes 17-25  model=gemini-3.1-pro-preview  finish=STOP  lines=9
tokens in/out/think=3220/18697/11788
2026-08-17 21:03  ne02 notes 1-13  model=gemini-3.1-pro-preview  finish=STOP  lines=13
tokens in/out/think=3232/28987/9000
2026-08-17 21:08  ne02 notes 14-25  model=gemini-3.1-pro-preview  finish=STOP  lines=12
tokens in/out/think=3248/24690/13287
2026-08-17 21:12  ne03 notes 1-13  model=gemini-3.1-pro-preview  finish=STOP  lines=13
tokens in/out/think=3242/25812/9541
2026-08-17 21:16  ne03 notes 14-25  model=gemini-3.1-pro-preview  finish=STOP  lines=12
tokens in/out/think=3244/29900/5026
2026-08-17 21:20  ne04 notes 1-13  model=gemini-3.1-pro-preview  finish=STOP  lines=13
tokens in/out/think=3264/24308/6754
2026-08-17 21:24  ne04 notes 14-25  model=gemini-3.1-pro-preview  finish=STOP  lines=12
tokens in/out/think=3266/23543/12904
2026-08-17 21:27  cs11 notes 19-25  model=gemini-3.1-pro-preview  finish=STOP  lines=7
tokens in/out/think=3249/14128/5363
```

## S5. Mechanical validation of generated batches

Every returned batch passed an automated battery before acceptance: JSONL validity; every annotation an exact substring of its note; zero Cyrillic characters; Greek-letter share; presence per note of patient name, AMKA, hospital number, date of birth, dates, address, telephone and clinician; AMKA consistency with the stated date of birth; telephone area code consistent with the hospital's city; identifier-label variety (at least three variants per field per batch); layout variety; absence of prohibited identifier types (AFM, ADT, AMA, EKAA); and no duplicate patient names within a batch. Identical full names recur across notes of different batches, as common Greek names do in reality; each note carries a distinct fictional identity (registry numbers, birth date, address). Batches failing materially were regenerated. Where a generated annotation did not match its note verbatim, it was repaired to the exact substring where the match was unambiguous (31 instances) or itemised for the clinician review pass; notes missing clinical-content annotations received a dedicated re-annotation request (23 notes). Three words written with Cyrillic look-alike characters were detected by the script check and corrected. All 500 notes were then re-verified. The full item list accompanies the review interface. After assembly, hospital website and e-mail contact lines were added deterministically to the letterheads of 141 notes, reflecting real-world practice (verified official domains; institutions such as the Onassis Cardiac Surgery Center print both on every page, public hospitals variably or not at all), with corresponding url/email annotations; one clinical reference URL (a drug-safety resource) is annotated as content that must survive. The generation prompts (S2–S3) were drafted, and the batch generation and the mechanical validation battery described in this section were executed, by a large language model assistant (Claude Fable 5, Anthropic) operating under the explicit instructions of the senior author (FPR), who specified the corpus design, reviewed the validation outputs at each step, and made all acceptance decisions.

## S6. TRIPOD-LLM checklist

Completed with the interactive TRIPOD-LLM checklist (https://tripod-llm.vercel.app/, accessed 18 September 2026; Gallifant et al., Nat Med 2025;31:60–69). Item numbers and wording are those of the website.

Scope. The system evaluated in this study is not a large language model (LLM). An LLM was used for one purpose: to write the synthetic corpus under a fixed prompt pack (S2–S4). The checklist is therefore completed for that part of the study, and items that presuppose an evaluated LLM are answered as not applicable, with the reason. The website requires a research design to be chosen; none of its four designs describes the generation of a dataset, and "LLM evaluation" with the task "Documentation generation" was chosen as the closest, because the generated notes were assessed by automated validation (S5) and by full review by two clinicians. Items marked "not required" are those the website excludes for this design and task.

*M = section of the main text; S = section of this supplement; — = not reported.*

| Item | Checklist item (TRIPOD-LLM) | Reported in | Note |
|---|---|---|---|
| **Title** | | | |
| 2a | Identify the study as developing, fine-tuning, and/or evaluating the performance of an LLM, specifying the task, the target population, and the outcome to be predicted. | Title | Partly applicable. The title names the evaluated engine and the benchmark; the engine is not an LLM, so the study is not presented as LLM development or evaluation. |
| **Abstract** | | | |
| 2b | Provide a brief explanation of the healthcare context, use case and rationale for developing or evaluating the performance of an LLM. | Abstract (Background and Objective) | Healthcare context and rationale are given for the de-identification task; the LLM is a corpus generator, not the system evaluated. |
| **Objectives** | | | |

| Item | Checklist item (TRIPOD-LLM) | Reported in | Note |
|---|---|---|---|
| 2c | Specify the study objectives, including whether the study describes LLMs development, tuning, and/or evaluation | Abstract (Background and Objective) | Objective stated. No LLM is developed, tuned or evaluated as the system under test. |
| **Methods** | | | |
| 2d | Describe the key elements of the study setting. | Abstract (Methods) | Setting: 500 synthetic Greek discharge notes processed by a fully local engine. |
| 2e | Detail all data used in the study, specify data splits and any selective use of data. | Abstract (Methods) | All 500 notes are evaluation data; there is no training or tuning split, and no data were used selectively. |
| 2f | Specify the name and version of LLM used. | Abstract (Methods) | Gemini 3.1 Pro (Google); exact model string in S4. |
| 2g | Briefly summarize the LLM-building steps, including any fine-tuning, reward modeling, reinforcement learning with human feedback (RLHF), etc. | — | Not required by the checklist for this research design and task. |
| 2h | Describe the specific tasks performed by the LLMs (e.g., medical QA, summarization, extraction), highlighting key inputs and outputs used in the final LLM. | Abstract (Methods) | Task of the LLM: writing synthetic discharge notes, with generation-time annotations, under fixed document layouts. |
| 2i | Specify the evaluation datasets/populations used, including the endpoint evaluated, and detail whether this information was held out during training/tuning where relevant, and what measure(s) were used to evaluate LLM performance. | Abstract (Methods) | Evaluation dataset and measures are stated (de-identification rate, over-redaction rate). The corpus was held out: no component of the engine was tuned against it (M 4). |
| **Results** | | | |
| 2j | Give an overall report and interpretation of the main results. | Abstract (Results) | Reported. |
| **Discussion** | | | |

| Item | Checklist item (TRIPOD-LLM) | Reported in | Note |
|---|---|---|---|
| 2k | Explicitly state any broader implications or concerns that have arisen in light of these results. | Abstract (Conclusions) | Reported. |
| **Other** | | | |
| 2l | Give the registration number and name of the registry or repository (if relevant). | — | Not required by the checklist for this research design and task. |
| **Background** | | | |
| 3a | Explain the healthcare context / use case (e.g., administrative, diagnostic, therapeutic, clinical workflow) and rationale for developing or evaluating the LLM, including references to existing approaches and models. | M 1; M 4 | Context, rationale and existing approaches (i2b2, Philter, synthetic corpora) are given for the de-identification task. |
| 3b | Describe the target population and the intended use of the LLM in the context of the care pathway, including its intended users in current gold standard practices (e.g., healthcare professionals, patients, public, or administrators). | M 1; M 2.1 | Intended users and setting of the evaluated engine are described. The LLM has no role in any care pathway: it was used once, offline, to write the corpus. |
| **Objectives** | | | |
| 4 | Specify the study objectives, including whether the study describes the initial development, fine-tuning, or validation of an LLM (or multiple stages). | M 1 (last paragraph) | Objective stated; no LLM development, fine-tuning or validation stage. |
| **Data** | | | |
| 5a | Describe the sources of data separately for the training, tuning, and/or evaluation datasets and the rationale for using these data (e.g., web corpora, clinical research/trial data, EHR data). | M 2.2; S2–S4 | No training or tuning data. The evaluation corpus was generated by the LLM under the prompt pack; rationale: no shareable Greek clinical corpus exists. |

| Item | Checklist item (TRIPOD-LLM) | Reported in | Note |
|---|---|---|---|
| 5b | Describe the relevant data points and provide a quantitative and qualitative description of their distribution and other relevant descriptors of the dataset (e.g., source, languages, countries of origin) | M 2.2–2.3; Supplementary Tables 2–3 | 500 notes in Greek with embedded English clinical terms; 300 cardiac surgery, 100 cardiology, 100 nephrology; identifier and probe counts by category. |
| 5c | Specifically state the date of the oldest and newest item of text used in the development process (training, fine-tuning, reward modeling) and in the evaluation datasets. | S4 | All text was generated on 17 August 2026 (request log). Hospital contact lines were added to 141 letterheads after assembly (S5). |
| 5d | Describe any data pre-processing and quality checking, including whether this was similar across text corpora, institutions, and relevant sociodemographic groups. | S5 | Automated validation battery on every batch, repairs itemised; then full review of every note by two clinicians (M 2.4). |
| 5e | Describe how missing and imbalanced data were handled and provide reasons for omitting any data. | S4; S5 | Truncated deliveries were completed by range-specific top-up requests; batches failing validation were regenerated; the 300/100/100 specialty mix was set by design. No note was omitted. |
| **Analytical Methods** | | | |
| 6a | Report the LLM name, version, and last date of training or use during inference. | M 2.2; S4; Declarations | gemini-3.1-pro-preview, used 17 August 2026. Claude Fable 5 (Anthropic) drafted the prompt pack. The vendors' training cut-off dates are not reported. |
| 6b | Specify the type of LLM architecture, and LLM building steps, including any hyperparameter tuning (e.g., temperature, length limits, penalties), prompt engineering, and any inference settings (e.g., seed, temperature, max token length) as relevant. | S4 | Proprietary hosted model (architecture not disclosed by the vendor); temperature 0.9; streaming API; two requests per batch of 25 notes; no seed was set. No hyperparameter tuning. |
| 6c | Report details of LLM development process | — | Not required by the checklist for this |

| Item | Checklist item (TRIPOD-LLM) | Reported in | Note |
|---|---|---|---|
| | from text input to outcome generation, such as training, fine-tuning procedures, and alignment strategy (e.g., reinforcement learning, direct preference optimization, etc.) and alignment goals (e.g., helpfulness, honesty, harmlessness, etc.). | | research design and task. |
| 6d | Specify the initial and post-processed output of the LLM (e.g., probabilities, classification, unstructured text). | S2; S5 | Initial output: JSONL, one object per note (text plus annotations). Post-processing: verbatim-substring repair of annotations and the additions itemised in S5. |
| 6e | Provide details and rationale for any classification and how the probabilities were determined and thresholds identified. | — | Not required by the checklist for this research design and task. |
| 6f | Include metrics that capture the quality of generative outputs, such as consistency, relevance, and accuracy, compared to gold standards. | S5; M 2.4; M 3.1 | No reference text exists for a generated note, so no overlap metric applies. Quality was assured by the validation battery and by two clinicians reviewing every note; clinical realism was not formally scored (see 19b). |
| 6g | Report the outcome metrics' relevance to downstream task at deployment time and correlation of metric to human evaluation of the text for the intended use. | M 2.5 | Partly applicable. The outcome measures concern the evaluated engine. For the corpus, the downstream relevance is whether results transfer to real notes, addressed as a limitation (M 4). |
| **LLM Output** | | | |
| 7a | Clearly define the outcome, how the LLM predictions were calculated (e.g., formula, code, object, API), and evaluation metrics. | M 2.5; S5 | Study outcomes (de-identification and over-redaction rates) are defined for the engine, which is not an LLM. For the LLM, the outcome is an accepted note; acceptance criteria are in S5. |

| Item | Checklist item (TRIPOD-LLM) | Reported in | Note |
|---|---|---|---|
| 7b | If outcome assessment requires subjective interpretation, describe the qualifications of the assessors, any instructions provided, relevant information on demographics of the assessors, and inter-assessor agreement. | M 2.4; M 3.1 | Two clinicians reviewed all notes independently and blinded to each other; agreement F1 = 0.998; arbitration by a senior author. Reviewers: a junior doctor and a resident doctor in cardiothoracic surgery; arbiter: a consultant cardiothoracic surgeon (M 2.4). Reviewer demographics are not reported. |
| 7c | Specify how performance was compared to other LLMs, humans, and other benchmarks or standards. | — | Not done. No other LLM was compared as a generator. The de-identification engine is positioned against published systems in M 4. |
| **Annotation** | | | |
| 8a | If annotation was done, report how text was labeled, including providing specific annotation guidelines with examples. | M 2.3–2.4; S2 | Identifier categories and labelling rules are defined in the prompt pack (S2) and Methods; annotation was machine pre-annotation corrected by reviewers in a purpose-built interface. A separate guideline with worked examples was not produced. |
| 8b | If annotation was done, report how many annotators labeled the dataset(s), including the proportion of data in each dataset that were annotated by more than 1 annotator. | M 2.4 | Two annotators; 100% of notes reviewed by both. |
| 8c | If annotation was done, provide information on the background and experience of the annotators, and the inter-annotator agreement. | M 2.4; M 3.1 | Both annotators are clinicians; inter-annotator agreement F1 = 0.998 (31 disagreements in 22 notes). Annotators: a junior doctor and a resident doctor in cardiothoracic surgery (M 2.4). |
| **Prompting** | | | |

| Item | Checklist item (TRIPOD-LLM) | Reported in | Note |
|---|---|---|---|
| 9a | If research involved prompting LLMs, provide details on the processes used during prompt design, curation, and selection. | M 2.2; S2–S3 | Prompts reproduced verbatim: one master prompt and one batch prompt per batch. Layouts and label variants were fixed in advance; the LLM contributed narrative and fictional identities only. |
| 9b | If research involved prompting LLMs, report what data were used to develop the prompts. | M 2.2 | Prompts were developed from the layouts and identifier labels of authentic Greek hospital documents known to the clinician author; no patient data were used. |
| **Summarization** | | | |
| 10 | Describe any preprocessing of the data before summarization. | — | Not required by the checklist for this research design and task. |
| **Instruction Tuning / Alignment** | | | |
| 11 | If instruction tuning/alignment strategies were used, what were the instructions and interface used for evaluation, and what were the characteristics of the populations doing evaluation? | — | Not required by the checklist for this research design and task. |
| **Compute** | | | |
| 12 | Report compute, or proxies thereof (e.g., time on what and how many machines, cost on what and how many machines, inference time, floating-point operations per second (FLOPs)), required to carry out methods. | S4; M 3.3 | Per-request input, output and reasoning token counts are logged; generation took one day. The evaluated engine's runtime and memory on a consumer laptop are in M 3.3. |
| **Ethics Approval** | | | |
| 13 | Name the institutional research board or ethics committee that approved the study and describe the participant-informed consent or the ethics committee waiver of informed consent. | Declarations (Ethics statement) | Stated: no patients, participants or patient data were involved; ethics committee approval and informed consent were not required. |

| Item | Checklist item (TRIPOD-LLM) | Reported in | Note |
|---|---|---|---|
| **Open Science** | | | |
| 14a | Give the source of funding and the role of the funders for the present study. | Declarations | No funding. |
| 14b | Declare any conflicts of interest and financial disclosures for all authors. | Declarations | Reported, including the authors' role in the development of the engine. |
| 14c | Indicate where the study protocol can be accessed or state that a protocol was not prepared. | — | Not required by the checklist for this research design and task. |
| 14d | Provide registration information for the study, including register name and registration number, or state that the study was not registered. | — | Not required by the checklist for this research design and task. |
| 14e | Provide details of the availability of the study data. | Data availability; S7 | Corpus, annotations, outputs and all scored decisions are public (Zenodo DOI, with GitHub and Hugging Face mirrors), CC BY 4.0. |
| 14f | Provide details of the availability of the code to reproduce the study results. | Data availability; S7 | Scoring code released (MIT). The engine's source code is proprietary and is not released; every reported number can be recomputed from the released outputs. |
| **Public Involvement** | | | |
| 15 | Provide details of any patient and public involvement during the design, conduct, reporting, interpretation, or dissemination of the study or state no involvement. | — | Not required by the checklist for this research design and task. |
| **Participants** | | | |
| 16a | When using patient/EHR data, describe the flow of text/EHR/patient data through the study, including the number of | — | Not applicable: no patient or EHR data. Flow of the synthetic notes: Supplementary Figure 1. |

| Item | Checklist item (TRIPOD-LLM) | Reported in | Note |
|---|---|---|---|
| | documents/questions/participants with and without the outcome/label and follow-up time. | | |
| 16b | When using patient/EHR data, report the characteristics overall and, for each data source or setting, and for development/evaluation splits, including the key dates, key predictors, and sample size. | — | Not applicable: no patient or EHR data. Corpus characteristics: Supplementary Tables 2–3. |
| 16c | For LLM evaluation, show a comparison of the distribution of important predictors between development and evaluation data. | — | Not applicable: there are no development data. |
| 16d | When using patient/EHR data, specify the number of participants and outcome events in each analysis (e.g., for LLM development, hyperparameter tuning, LLM evaluation). | — | Not applicable: no participants. Number of notes and identifier instances in each analysis: M 3.2, Table 1. |
| **Performance** | | | |
| 17 | Report LLM performance according to pre-specified metrics (see item 7a) and/or human evaluation (see item 7d). | S5; M 3.1 | For the LLM: 31 annotations repaired, 23 notes re-annotated, three script corrections, and the reviewers' corrections summarised by the agreement figures. The study's performance results (M 3.2–3.4) are those of the engine. |
| **LLM Updating** | | | |
| 18 | If applicable, report the results from any LLM updating, including the updated LLM and subsequent performance. | — | Not applicable: no LLM was updated. |
| **Interpretation** | | | |
| 19a | Give an overall interpretation of the main results, including issues of fairness in the context of the objectives and previous | M 4 | Main results interpreted against prior work. The make-up of the fictional patients is given in Supplementary Table 4; all were |

| Item | Checklist item (TRIPOD-LLM) | Reported in | Note |
|---|---|---|---|
| | studies. | | adults and all but one carried a name in Greek script, which is stated as a limitation (M 4). Performance was not stratified by patient characteristics. |
| **Limitations** | | | |
| 19b | Discuss any limitations of the study and their effects on any biases, statistical uncertainty, and generalizability. | M 4 (Limitations) | Synthetic corpus may flatter real-world performance; generator habits; arbitration by an author involved in the engine's development; single machine. |
| **Usability of the LLM in context** | | | |
| 19c | Describe any known challenges in using data for the specified task and domain context with reference to representation, missingness, harmonization, and bias. | M 4 (Limitations) | Representation: a generator's habits may not span real documentation styles. |
| 19d | Define the intended use for the implementation under evaluation, including the intended input, end-user, level of autonomy/human oversight. | M 2.1; S1 | Intended use of the evaluated engine, its input, end-user and the human review of flagged documents are described. The LLM's only use was corpus generation under the authors' supervision. |
| 19e | If applicable, describe how poor quality or unavailable input data should be assessed and handled when implementing the LLM, i.e., what is the usability of the LLM in the context of current clinical care. | M 2.1; S1 | Partly applicable. Handling of poor input (scanned documents, character repair) is described for the engine; not applicable to the LLM. |
| 19f | If applicable, specify whether users will be required to interact in the handling of the input data or use of the LLM, and what level of expertise is required of users. | M 2.1 | The engine's user reviews the outgoing text of every flagged document before it leaves the machine; no LLM interaction is required of users in this study. |
| 19g | Discuss any next steps for future research, with a specific view to applicability and | M 4 (Limitations) | Real-note validation at Greek hospitals; testing on other machines and operating |

| Item | Checklist item (TRIPOD-LLM) | Reported in | Note |
|---|---|---|---|
| | generalizability of the LLM. | | systems. |

## S7. Public release of the benchmark (data and code availability)

The complete benchmark is publicly released as a companion package (Zenodo: https://doi.org/10.5281/zenodo.22824895; mirrored on GitHub: https://github.com/prorris/greek-deid-benchmark and on the Hugging Face Hub, prorris/greek-deid-benchmark). The package contains: (i) the corpus — all 500 synthetic notes as JSONL (full corpus and per-specialty files) together with their realistic hospital-layout PDF renderings; (ii) the annotations — the arbitrated gold standard of 7,937 must-redact identifier spans, both reviewers' complete independent corrected annotation sets, the inter-annotator agreement report and disagreement list, and the arbitration decisions; (iii) the scoring implementation — the exact scripts that compute every outcome reported in this study (outcome definitions, the attributable-part rule, confidence intervals, per-category and per-specialty breakdowns) and the run-to-run reproducibility comparison, together with the run harnesses documenting how the outgoing text was produced; and (iv) the results — the full de-identified output text of every note in both independent benchmark runs (pipeline fingerprint 765328fbf1241ea4), the egress-review flags, per-note processing times, and all scored decisions of both runs. Every number reported in this paper can therefore be recomputed from the released materials, and any other de-identification system can be evaluated against the same corpus, gold standard and scoring rules. All released content is synthetic; no real patient data was used at any stage. Data are released under the CC BY 4.0 license and the scoring scripts under the MIT license.

The de-identification engine itself is proprietary and is identified by its published pipeline fingerprint.

## S8. Supplementary tables and figures

**Supplementary Table 1. Third-party software components of the evaluated engine (Morpheus 1.0, pipeline fingerprint 765328fbf1241ea4).**

| Component | Role in the evaluated engine | Version or revision |
|---|---|---|
| Microsoft Presidio (presidio-analyzer, presidio-anonymizer) | Recogniser framework of the English-language layer: pattern rules and model recognisers, scored spans | 2.2.359, 2.2.362 |
| spaCy with the en_core_web_lg model | English language model (tokenisation and named entities) | spaCy 3.8.13; en_core_web_lg 3.8.0 |
| obi/deid_roberta_i2b2 | Clinical de-identification transformer: RoBERTa fine-tuned on the i2b2 2014 de-identification corpus | model revision 78f2152eb93ddd817290ce8dbe46f1a6685e09fc |
| GLiNER with the urchade/gliner_multi-v2.1 model | Zero-shot named-entity recognition: Greek person and location candidates | gliner 0.2.26; model revision 443d26d654e0324125a96bebd8e796c14ff2efe6 |
| Hugging Face transformers; PyTorch | Model runtime (CPU inference, four threads pinned) | transformers 5.1.0; torch 2.12.0 |
| pdfplumber; Tesseract OCR (through pytesseract) | Document reading in the application: embedded PDF text, and optical character recognition of scans. Not exercised in this study, in which notes were supplied as text | pdfplumber 0.11.9; pytesseract 0.3.13 |

*Versions and model revisions are those recorded in the frozen configuration of the evaluated engine, which the pipeline fingerprint summarises together with the checksums*

*of the engine's vocabulary files and its run settings. The Greek-specific rules, vocabularies and gazetteers, the surrogate replacement, the tokeniser and the egress guard are the authors' own and are not third-party components.*

**Supplementary Table 2. Annotated must-redact identifiers in the benchmark corpus.**

| Identifier | Instances | % of all | Notes containing, n (%) |
|---|---|---|---|
| **Direct identifiers** | **4,960** | **62.5** | |
| Patient name | 775 | 9.8 | 500 (100) |
| Address | 1,655 | 20.9 | 500 (100) |
| Telephone number | 1,047 | 13.2 | 500 (100) |
| Date of birth | 500 | 6.3 | 500 (100) |
| Social security number (AMKA) | 500 | 6.3 | 500 (100) |
| Identity card / hospital number | 483 | 6.1 | 483 (96.6) |
| **Quasi-identifiers** | **2,977** | **37.5** | |
| Dates of care | 1,686 | 21.2 | 500 (100) |
| Hospital name | 521 | 6.6 | 500 (100) |
| Clinician name | 500 | 6.3 | 500 (100) |
| Hospital website (URL) | 141 | 1.8 | 141 (28.2) |
| Hospital email address | 129 | 1.6 | 129 (25.8) |
| **Total** | **7,937** | **100.0** | **500 (100)** |

*Five hundred synthetic Greek discharge notes; 7,937 identifier instances. Percentages are of all must-redact instances. Address and telephone comprise both the patient's own and the institution's values. Counts are from the arbitrated gold standard (dual independent full review with arbitration of all disagreements).*

**Supplementary Table 3. Must-survive clinical probes in the benchmark corpus.**

| Clinical probe category | Instances | % of all | Notes containing, n (%) |
|---|---|---|---|
| Drug names | 1,849 | 27.4 | 488 (97.6) |
| Diagnoses | 993 | 14.7 | 491 (98.2) |
| Laboratory values | 953 | 14.1 | 450 (90.0) |
| Section headings | 835 | 12.4 | 463 (92.6) |
| Procedures | 737 | 10.9 | 436 (87.2) |
| Measurements | 649 | 9.6 | 375 (75.0) |
| Clinical abbreviations | 554 | 8.2 | 366 (73.2) |
| Eponyms | 177 | 2.6 | 161 (32.2) |
| **Total** | **6,747** | **100.0** | **500 (100)** |

*Six thousand seven hundred and forty-seven probe instances across the same 500 notes. Percentages are of all must-survive instances. Counts are from the corpus as accepted; the dual review and arbitration concerned must-redact identifiers and did not modify must-survive probes.*

**Supplementary Table 4. Characteristics of the fictional patients in the benchmark corpus.**

| Characteristic | All notes (n = 500) | Cardiac surgery (n = 300) | Cardiology (n = 100) | Nephrology (n = 100) |
|---|---|---|---|---|
| Male, n (%) | 314 (62.8) | 195 (65.0) | 60 (60.0) | 59 (59.0) |
| Female, n (%) | 186 (37.2) | 105 (35.0) | 40 (40.0) | 41 (41.0) |
| Age in years, median (IQR) | 63 (53–72) | 65 (56–72) | 64 (53–72) | 58 (44–67) |
| Age in years, range | 21–92 | 22–92 | 22–89 | 21–88 |

| Characteristic | All notes (n = 500) | Cardiac surgery (n = 300) | Cardiology (n = 100) | Nephrology (n = 100) |
|---|---|---|---|---|
| Aged 18–39, n | 59 | | | |
| Aged 40–64, n | 214 | | | |
| Aged 65–79, n | 193 | | | |
| Aged 80 and over, n | 34 | | | |
| Aged under 18, n | 0 | | | |
| Patient name written in Latin script, n | 1 | | | |

*Derived from the released corpus. Age is the interval between the annotated date of birth and the first annotated date of care in the note (all 500 notes). Sex was read from the grammatical gender the note itself uses for the patient (491 notes) and otherwise from the patient's given name (9 notes); where both were available (477 notes) they agreed in every case. No note states ethnicity or nationality; one patient, in a note from an international patients' department, has a name in Latin script. No note concerns a patient under 18 years of age.*

Supplementary Figure 1. Corpus construction, annotation and benchmark flow. Box values are counts from the study record; the two runs of the frozen pipeline are compared decision-by-decision (Reproducibility, main text).

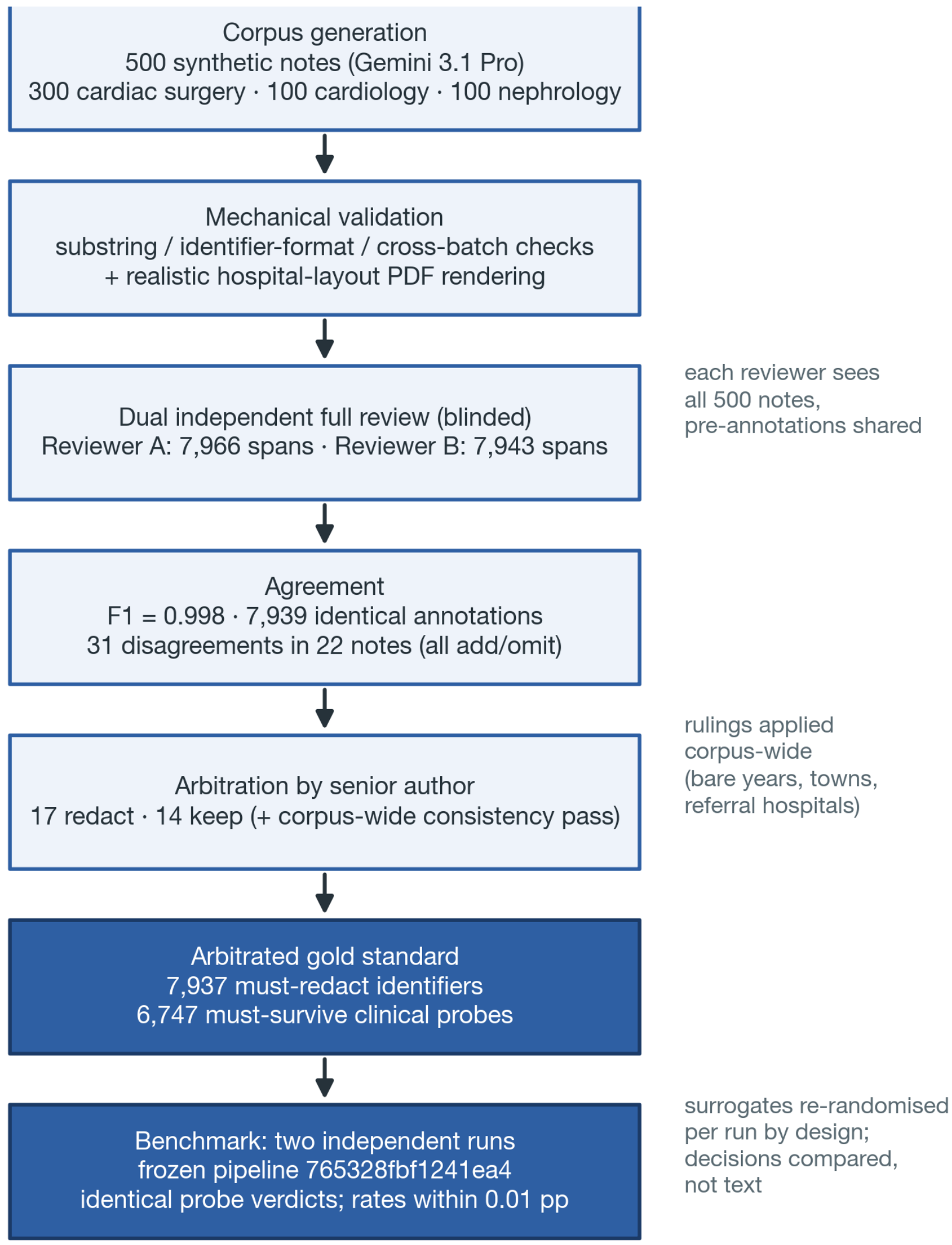
Corpus generation
500 synthetic notes (Gemini 3.1 Pro)
300 cardiac surgery · 100 cardiology · 100 nephrology
Mechanical validation
substring / identifier-format / cross-batch checks
+ realistic hospital-layout PDF rendering
Dual independent full review (blinded)
Reviewer A: 7,966 spans · Reviewer B: 7,943 spans
each reviewer sees all 500 notes, pre-annotations shared
Agreement
F1 = 0.998 · 7,939 identical annotations
31 disagreements in 22 notes (all add/omit)
Arbitration by senior author
17 redact · 14 keep (+ corpus-wide consistency pass)
rulings applied corpus-wide (bare years, towns, referral hospitals)
Arbitrated gold standard
7,937 must-redact identifiers
6,747 must-survive clinical probes
Benchmark: two independent runs
frozen pipeline 765328fbf1241ea4
identical probe verdicts; rates within 0.01 pp
surrogates re-randomised per run by design; decisions compared, not text